\documentclass{article}

\usepackage{microtype}
\usepackage{graphicx}
\usepackage{subcaption}
\usepackage{booktabs} 
\usepackage{enumitem}
\usepackage{dsfont}         
\usepackage{amssymb}        
\usepackage{placeins}

\usepackage[nomargin,inline,draft]{fixme}
\fxsetup{theme=color,mode=multiuser}

\usepackage{hyperref}

\newcommand{\rmm}{\mathbf{m}}

\newcommand{\spm}[1]{{\scriptstyle \pm {#1}}}
\usepackage[dvipsnames]{xcolor}

\usepackage[accepted]{icml2022}

\usepackage{amsmath,amsfonts,bm}

\def\eqref#1{equation~\ref{#1}}

\def\1{\bm{1}}

\def\rvh{{\mathbf{h}}}

\def\rvp{{\mathbf{p}}}

\def\rvx{{\mathbf{x}}}

\def\rmI{{\mathbf{I}}}

\def\rmQ{{\mathbf{Q}}}
\def\rmR{{\mathbf{R}}}

\def\vmu{{\bm{\mu}}}

\def\vepsilon{{\bm{\epsilon}}}

\def\vh{{\bm{h}}}

\def\vt{{\bm{t}}}
\def\vu{{\bm{u}}}

\def\vx{{\bm{x}}}

\def\vz{{\bm{z}}}

\DeclareMathAlphabet{\mathsfit}{\encodingdefault}{\sfdefault}{m}{sl}
\SetMathAlphabet{\mathsfit}{bold}{\encodingdefault}{\sfdefault}{bx}{n}

\newcommand{\R}{\mathbb{R}}

\icmltitlerunning{Equivariant Diffusion for Molecule Generation in 3D}

\begin{document}

\twocolumn[
\icmltitle{Equivariant Diffusion for Molecule Generation in 3D}



\icmlsetsymbol{equal}{*}

\begin{icmlauthorlist}
\icmlauthor{Emiel Hoogeboom}{equal,delta}
\icmlauthor{Victor Garcia Satorras}{equal,delta}
\icmlauthor{Cl\'ement Vignac}{equal,epfl}
\icmlauthor{Max Welling}{delta}
\end{icmlauthorlist}

\icmlaffiliation{delta}{UvA-Bosch Delta Lab, University of Amsterdam, Netherlands}
\icmlaffiliation{epfl}{EPFL, Lausanne, Switzerland}

\icmlcorrespondingauthor{Emiel Hoogeboom}{e.hoogeboom@uva.nl}
\icmlcorrespondingauthor{Victor Garcia Satorras}{v.garciasatorras@uva.nl}
\icmlcorrespondingauthor{Cl\'ement Vignac}{clement.vignac@epfl.ch}

\icmlkeywords{Denoising, Diffusion, Molecule, Generation, 3D}

\vskip 0.3in
]



\printAffiliationsAndNotice{\icmlEqualContribution} 

\begin{abstract} 
This work introduces a diffusion model for molecule generation in 3D that is equivariant to Euclidean transformations. Our E(3) Equivariant Diffusion Model (EDM) learns to denoise a diffusion process with an equivariant network that jointly operates on both continuous (atom coordinates) and categorical features (atom types). In addition, we provide a probabilistic analysis which admits likelihood computation of molecules using our model. Experimentally, the proposed method significantly outperforms previous 3D molecular generative methods regarding the quality of generated samples and efficiency at training time.

\end{abstract}

\section{Introduction}

\begin{figure}[t!]
\vspace{-5pt}
\includegraphics[width=.48\textwidth]{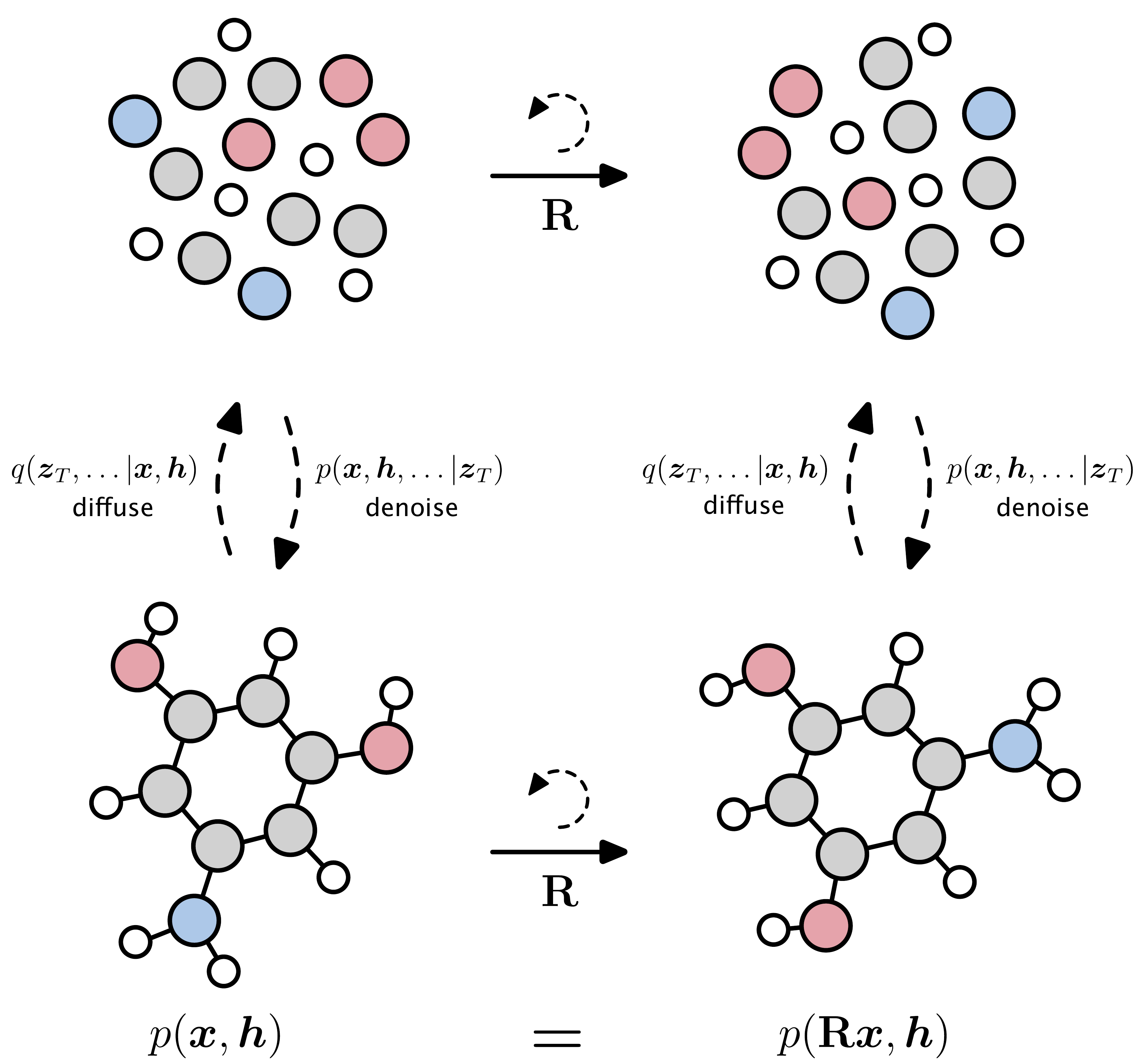}
\centering
\vspace{-18pt}
\caption{Overview of the EDM. To generate a molecule, a normal distributed set of points is denoised into a molecule consisting of atom coordinates $\vx$ in 3D and atom types $\vh$. 
As the model is rotation equivariant, the likelihood is preserved when a molecule is rotated by $\rmR$.}
\label{fig:banner}
\vspace{-8pt}
\end{figure}

Modern deep learning methods are starting to make an important impact on molecular sciences. 
Behind the success of Alphafold in protein folding prediction \citep{jumper2021highly}, an increasing body of literature develops deep learning models to analyze or synthesize (in silico) molecules \citep{simonovsky2018graphvae,gebauer2019symmetry,klicpera2020dimenet,simm2021symmetryaware}. 
 
Molecules live in the physical 3D space, and as such are subject to geometric symmetries such as translations, rotations, and possibly reflections. These symmetries are referred to as the Euclidean group in 3 dimensions, E(3).
Leveraging these symmetries in molecular data is important for good generalization and has been extensively studied \citep{thomas2018tensorfield,fuchs2020se3transformer,finzi2020lieconv}.

Although developed for discriminative tasks, E(n) equivariant layers can also be used for molecule generation in 3D. In particular, they have been integrated into  autoregressive models \citep{gebauer2018generating, gebauer2019symmetry} which artificially introduce an order in the atoms and are known to be difficult to scale during sampling \citep{xu2021anytimesampling}. Alternatively, continuous-time normalizing flows such as \citet{kohler2020equivariantflows} or E-NF \citep{satorras2021en_flows} are expensive to train since they have to integrate a differential equation, leading to limited performance and scalability.

In this work, we introduce E(3) Equivariant Diffusion Models (EDMs). EDMs learn to denoise a diffusion process that operates on both continuous coordinates and categorical atom types. To the best of our knowledge, it is the first diffusion model that directly generates molecules in 3D space. Our method does not require a particular atom ordering (in contrast to autoregressive models) and can be trained much more efficiently than normalizing flows. To give an example, EDMs generate up to 16 times more stable molecules than E-NFs when trained on QM9, while requiring half of the training time. This favourable scaling behaviour allows EDMs to be trained on larger drug-like datasets such as GEOM-Drugs \citep{axelrod2020geom}.

Our contributions can be summarized as follows. We introduce an equivariant denoising diffusion model that operates on atom coordinates and categorical features. We add a probabilistic analysis which allows likelihood computation and this analysis is consistent with continuous and categorical features. We show that our method outperforms previous molecule generation models in log-likelihood and molecule stability.

\section{Background}

\begin{figure*}[t]
\vspace{-2pt}
\includegraphics[width=.99\textwidth]{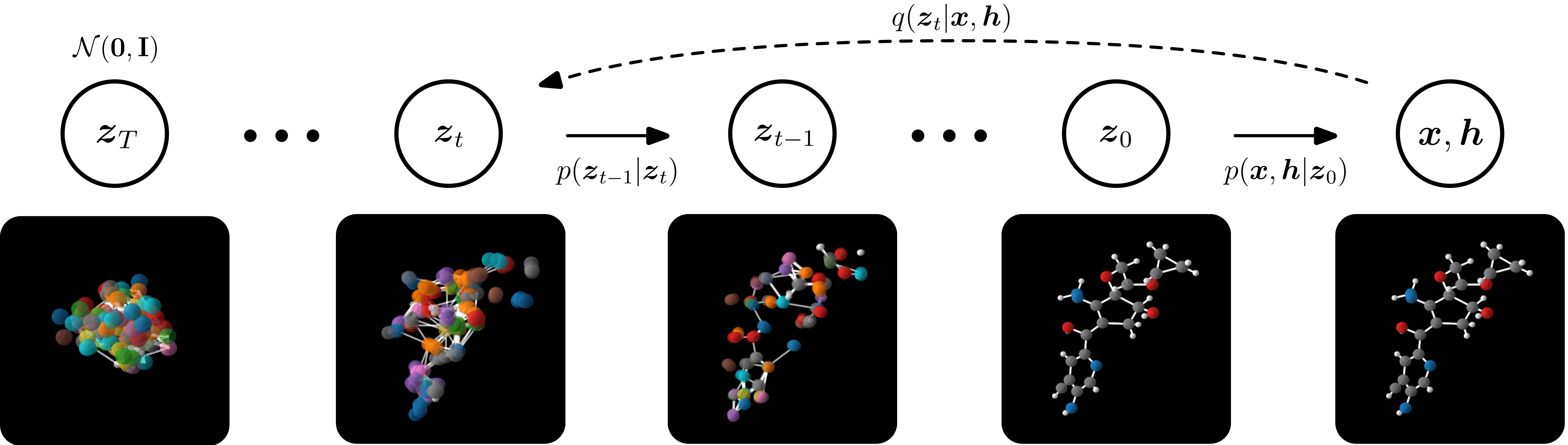}
\centering
\vspace{-7pt}
\caption{Overview of the Equivariant Diffusion Model. To generate molecules, coordinates $\vx$ and features $\vh$ are generated by denoising variables $\vz_t$ starting from standard normal noise $\vz_T$. This is achieved by sampling from the distributions $p(\vz_{t-1} | \vz_t)$ iteratively. To train the model, noise is added to a datapoint $\vx, \vh$ using $q(\vz_t | \vx, \vh)$ for the step $t$ of interest, which the network then learns to denoise.}
\label{fig:molecule_samples}
\vspace{-8pt}
\end{figure*}

\subsection{Diffusion Models}
Diffusion models learn distributions by modelling the reverse of a diffusion process: a \textit{denoising} process. Given a data point $\vx$, a diffusion process that adds noise to $\vz_t$ for $t = 0, \ldots, T$ is defined by the multivariate normal distribution:
\begin{equation} \label{eq:noising_process}
    q(\vz_t | \vx) = \mathcal{N}(\vz_t | \alpha_t \vx_t, \sigma_t^2 \rmI),
\end{equation}
where $\alpha_t \in \R^+$ controls how much signal is retained and $\sigma_t \in \R^+$ controls how much noise is added. In general, $\alpha_t$ is modelled by a function that smoothly transitions from $\alpha_0 \approx 1$ towards $\alpha_T \approx 0$.
A special case of noising process is the variance preserving process \citep{sohldickstein2015diffusion,ho2020denoising} for which $\alpha_t = \sqrt{1 - \sigma_t^2}$. Following \citet{kingma2021variational}, we define the signal to noise ratio $\mathrm{SNR}(t) = \alpha_t^2 / \sigma_t^2$, which simplifies notations.
This diffusion process is Markov and can be equivalently written with transition distributions as:
\begin{equation}
    q(\vz_t | \vz_s) = \mathcal{N}(\vz_t | \alpha_{t|s}\vz_s, ~\sigma_{t|s}^2 \rmI),
\end{equation}
for any $t > s$ with $\alpha_{t|s} = \alpha_t / \alpha_s$ and $\sigma_{t|s}^2 = \sigma_t^2 - \alpha_{t|s}^2 \sigma_s^2$. The entire noising process is then written as:
\begin{equation}
    q(\vz_0, \vz_1, \ldots, \vz_T | \vx) = q(\vz_0 | \vx) \prod\nolimits_{t=1}^T q(\vz_t | \vz_{t-1}).             
\end{equation}
The posterior of the transitions conditioned on $\vx$ gives the inverse of the noising process, the \textit{true denoising process}. It is also normal and given by:
\begin{equation}
    q(\vz_s | \vx, \vz_t) = \mathcal{N}(\vz_s | \vmu_{t \rightarrow s}(\vx, \vz_t), \sigma_{t \rightarrow s}^2 \rmI),
    \label{eq:background_noise_posterior}
\end{equation}
where the definitions for $\vmu_{t \rightarrow s}(\vx, \vz_t)$ and $\sigma_{t \rightarrow s}$ can be analytically obtained as
\small $$\vmu_{t \rightarrow s}(\vx, \vz_t) = \frac{\alpha_{t|s} \sigma_s^2}{\sigma_t^2} \vz_t + \frac{\alpha_s \sigma_{t|s}^2}{\sigma_t^2} \vx \quad\text{and}\quad \sigma_{t \rightarrow s} = \frac{\sigma_{t|s} \sigma_s}{\sigma_t}.$$ \normalsize
\paragraph{The Generative Denoising Process}
In contrast to other generative models, in diffusion models, the generative process is defined with respect to the \textit{true denoising process}. The variable $\vx$, which is unknown to the generative process, is replaced by an approximation $\hat{\vx} = \phi(\vz_t, t)$ given by a neural network $\phi$. Then the generative transition distribution $p(\vz_s | \vz_t)$ is chosen to be $q(\vz_s | \hat{\vx}(\vz_t, t), \vz_t)$. 
Similarly to Eq.\ \ref{eq:background_noise_posterior}, it can be expressed using the approximation $\hat\vx$ as:
\begin{equation}
    p(\vz_s | \vz_t) = \mathcal{N}(\vz_s | \vmu_{t \rightarrow s}(\hat{\vx}, \vz_t), \sigma_{t \rightarrow s}^2 \rmI).
    \label{eq:background_generative}
\end{equation}
With the choice $s=t-1$, a variational lower bound on the log-likelihood of $\vx$ given the generative model is given by:
\begin{equation}
    \log p(x) \geq \mathcal{L}_0 + \mathcal{L}_{\text{base}} + \sum\nolimits_{t=1}^{T} \mathcal{L}_t,
\end{equation}
where $\mathcal{L}_0 = \log p(\vx | \vz_0)$ models the likelihood of the data given $\vz_0$, $\mathcal{L}_{\text{base}} = -\mathrm{KL}(q(\vz_T | \vx) | p(\vz_T))$ models the distance between a standard normal distribution and the final latent variable $q(\vz_T | \vx)$, and
\begin{equation*}
    \mathcal{L}_t = -\mathrm{KL}(q(\vz_s | \vx, \vz_t) | p(\vz_s | \vz_t)) \quad \text{ for } t = 1, \ldots, T.
\end{equation*}
While in this formulation the neural network directly predicts $\hat{\vx}$, \citet{ho2020denoising} found that optimization is easier when predicting the Gaussian noise instead. Intuitively, the network is trying to predict which part of the observation $\vz_t$ is noise originating from the diffusion process, and which part corresponds to the underlying data point $\vx$. Specifically, if $\vz_t = \alpha_t \vx + \sigma_t \vepsilon$, then the neural network $\phi$ outputs $\hat{\vepsilon} = \phi(\vz_t, t)$, so that:
\begin{equation}
    \hat{\vx} = (1/ \alpha_t) ~\vz_t  - (\sigma_t / \alpha_t) ~ \hat{\vepsilon}
\end{equation} 
As shown in \citep{kingma2021variational}, with this parametrization $\mathcal L_t$  simplifies to:
\begin{equation}
\small
    \mathcal{L}_t = \mathbb{E}_{\vepsilon \sim \mathcal{N}(\mathbf{0}, \rmI)} \Big{[}\frac{1}{2}(1 - \mathrm{SNR}(t-1)/\mathrm{SNR}(t)) || \vepsilon - \hat{\vepsilon} ||^2\Big{]}
    \label{eq:loss_t_epsilon_param}
\end{equation}

In practice the term $\mathcal{L}_{\text{base}}$ is close to zero when the noising schedule is defined in such a way that $\alpha_T \approx 0$.
Furthermore, if $\alpha_0 \approx 1$ \textit{and} $\vx$ is discrete, then $\mathcal{L}_0$ is close to zero as well.

\subsection{Equivariance} \label{sec:equivariance}
A function $f$ is said to be equivariant to the action of a group $G$ if $T_g (f(\vx)) =  f(S_g(\vx))$ for all $g \in G$, where $S_g, T_g$ are linear representations related to the group element $g$ \citep{serre1977linear}.
In this work, we consider the Euclidean group $E(3)$ generated by translations, rotations and reflections, for which $S_g$ and $T_g$ can be represented by a translation $\vt$ and an orthogonal matrix $\rmR$ that rotates or reflects coordinates. $f$ is then equivariant to a rotation or reflection $\rmR$ if transforming its input results in an equivalent transformation of its output, or $\rmR f(\vx) =  f(\rmR\vx)$.

\paragraph{Equivariant Distributions and Diffusion}
In our setting, a conditional distribution $p(y | x)$ is equivariant to the action of rotations and reflections when
\begin{equation} \label{eq:equivariant_distribution}
    p(y | x) = p(\rmR y | \rmR x) \quad \text{ for all orthogonal } \rmR.
\end{equation} 
A distribution is invariant to $\mathbf{R}$ transformations if
\begin{equation}
p(y) = p(\rmR y) \quad  \text{ for all orthogonal } \rmR. 
\end{equation}
\citet{kohler2020equivariantflows} showed that an invariant distribution composed with an equivariant invertible function results in an invariant distribution. Furthermore, \citet{xu2022geodiff} proved that if $x \sim p(x)$ is invariant to a group and the transition probabilities of a Markov chain $y \sim p(y|x)$ are equivariant, then the marginal distribution of $y$ at any time step is invariant to group transformations as well.
This is helpful as it means that if $p(\vz_T)$ is invariant and
the neural network used to parametrize $p(\vz_{t-1} | \vz_t)$ is equivariant, 
then the marginal distribution $p(\vx)$ of the denoising model will be an invariant distribution as desired.



\paragraph{Points and Features in E(3)}
In this paper, we consider point clouds $\vx = (\vx_1, \dots, \vx_M )\in \mathbb{R}^{M \times 3}$ with corresponding features $\vh = (\vh_1, \dots, \vh_M )\in \mathbb{R}^{M \times \text{nf}}$. The features $\vh$ are invariant to group transformations, and the positions are affected by rotations, reflections and translations as $\rmR \vx + \vt = (\rmR\vx_1 + \vt, \dots, \rmR\vx_M + \vt)$ where $\rmR$ is an orthogonal matrix\footnote{As a matrix-multiplication the left-hand side would be written $\vx \rmR^T$. Formally $\rmR\vx$ can be seen as a group action of $\rmR$ on $\vx$.}.  The function $(\vz_x, \vz_h) = f(\vx, \vh)$ is $E(3)$ equivariant if for all orthogonal $\rmR$ and $\vt \in \R^3$ we have:
\begin{equation} \label{eq:equivariance}
\mathbf{R} \vz_x + \vt, \vz_h = f(\mathbf{R}\vx + \vt, \rvh)
\end{equation}
\textbf{E(n) Equivariant Graph Neural Networks (EGNNs)} \citep{satorras2021egnn}
are a type of Graph Neural Network that satisfies the equivariance constraint (\ref{eq:equivariance}). 
In this work, we consider interactions between all atoms, and therefore assume a fully connected graph $\mathcal G$ with nodes $v_i \in \mathcal{V}$.
Each node $v_i$ is endowed with coordinates $\vx_i \in \mathbb R^3$ as well as features $\vh_i \in \mathbb R^d$. 
In this setting, EGNN consists of the composition of Equivariant Convolutional Layers $\vx^{l+1}, \vh^{l+1} = \mathrm{EGCL}[\vx^l, \rvh^l]$ which are defined as:
\begin{align}
\rmm_{ij} &= \phi_{e}\left(\vh_{i}^{l}, \vh_{j}^{l}, d_{ij}^2, a_{ij}\right), \, \vh_{i}^{l+1} = \phi_{h}(\vh_{i}^l, { \sum_{j \neq i}} \tilde{e}_{ij} \rmm_{ij}),  \nonumber \\
\vx_{i}^{l+1} &= \vx_{i}^{l}+\sum_{j \neq i} \frac{\vx_{i}^{l}-\vx_{j}^{l}}{d_{ij}+ 1} \phi_{x}\left(\vh_{i}^{l}, \vh_{j}^{l}, d_{ij}^2, a_{ij}\right),
\label{eq:coord_update} 
\end{align}
where $l$ indexes the layer, and $d_{ij} = \| \vx_{i}^{l}-\vx_{j}^{l} \|_2$ is the euclidean distance between nodes $(v_i,v_j)$, and $a_{ij}$ are optional edge attributes. The difference ($\vx_i^l - \vx_j^l $) in Equation \ref{eq:coord_update} is normalized by $d_{ij} + 1$ as done in \citep{satorras2021en_flows} for improved stability, as well as the attention mechanism which infers a soft estimation of the edges $\tilde{e}_{ij}=\phi_{inf}(\rmm_{ij})$. All learnable components ($\phi_e$, $\phi_h$, $\phi_x$ and $\phi_{inf}$) are parametrized by fully connected neural networks (cf. Appendix~\ref{ap:the_dynamics} for details). An entire EGNN architecture is then composed of $L$ $\mathrm{EGCL}$ layers which applies the following non-linear transformation $\hat\vx, \hat\vh = \mathrm{EGNN}[\vx^0, \vh^0]$. This transformation satisfies the required equivariant property in Equation~\ref{eq:equivariance}.


\section{EDM: E(3) Equivariant Diffusion Model} \label{sec:method}
In this section we describe EDM, an E(3) Equivariant Diffusion Model. EDM defines a noising process on both node positions and features, and learns the generative \textit{denoising} process using an equivariant neural network. We also determine the equations for log-likelihood computation.

\subsection{The Diffusion Process}
We first define an equivariant diffusion process for coordinates $\vx_i$ with atom features $\vh_i$ that adds noise to the data.
Recall that we consider a set of points $\{(\vx_i, \vh_i)\}_{i=1, \ldots, M}$, where each node has associated to it a coordinate representation $\vx_i \in \mathbb{R}^{n}$ and an attribute vector $\vh_i \in \mathbb{R}^{\text{nf}}$. Let $[\cdot, \cdot]$ denote a concatenation. We define the equivariant noising process on latent variables $\vz_t = [\vz_t^{(x)}, \vz_t^{(h)}]$ as:
\begin{align}
\begin{split}
 q(\vz_t | \vx, \vh) = \mathcal{N}_{xh}(\vz_t | \alpha_t [\vx, \vh], \sigma_t^2 \rmI)
 \\
\end{split}
\label{eq:diffusion_points_features}
\end{align}
for $t=1, \ldots, T$ where $\mathcal{N}_{xh}$ is concise notation for the product of two distributions, one for the noised coordinates $\mathcal{N}_{x}$ and another for the noised features $\mathcal{N}$ given by:
\begin{equation}
    \mathcal{N}_{x}(\vz_t^{(x)} | \alpha_t \vx, \sigma_t^2 \rmI) \cdot \mathcal{N}(\vz_t^{(h)} | \alpha_t \vh, \sigma_t^2 \rmI)
\end{equation}
These equations correspond to Equation~\ref{eq:noising_process} in a standard diffusion model. Also, a slight abuse of notation is used to aid readability: technically $\vx$, $\vh$, $\vz_t$ are two-dimensional variables with an axis for the point identifier and an axis for the features. 
However, in the distributions they are treated as if flattened to a vector.

As explained in \citep{satorras2021en_flows} it is impossible to have a non-zero distribution that is invariant to translations, since it cannot integrate to one. However, one can use distributions on the linear subspace where the center of gravity is always zero. Following \citep{xu2022geodiff} that showed that such a linear subspace can be used consistently in diffusion, $\mathcal{N}_x$ is defined as a normal distribution on the subspace defined by $\sum_i \vx_i = \mathbf{0}$ for which the precise definition is given in Appendix~\ref{sec:center_gravity_normal}.

Since the features $\vh$ are invariant to E(n) transformations, the noise distribution for these features can be the conventional normal distribution $\mathcal{N}$. Although similar to standard diffusion models, depending on whether data is categorical (for the atom type), ordinal (for atom charge), or continuous, different starting representations of $\vh$ may be desirable and require different treatment in $\mathcal{L}_0$, on which we will expand in Section~\ref{sec:zeroth_likelihood}.

\begin{algorithm}[t]
   \caption{Optimizing EDM}
   \label{alg:optimize_edm}
\begin{algorithmic}
\STATE {\bfseries Input:} Data point $\vx$, neural network $\phi$
\STATE Sample $t \sim \mathcal{U}(0, \ldots, T)$, $\vepsilon \sim \mathcal{N}(\mathbf{0}, \rmI)$
\STATE Subtract center of gravity from $\vepsilon^{(x)}$ in $\vepsilon = [\vepsilon^{(x)}, \vepsilon^{(h)}]$
\STATE Compute $\vz_t = \alpha_t [\vx, \vh] + \sigma_t \epsilon$
\STATE Minimize $||\vepsilon - \phi(\vz_t, t)||^2$
\end{algorithmic}
\end{algorithm}

\begin{algorithm}[t]
   \caption{Sampling from EDM}
   \label{alg:sample_edm}
\begin{algorithmic}
\STATE Sample $\vz_T \sim \mathcal{N}(\mathbf{0}, \rmI)$
\FOR{$t$ in $T,\, T-1, \ldots, 1$ where $s = t - 1$}
\STATE Sample $\vepsilon \sim \mathcal{N}(\mathbf{0}, \rmI)$
\STATE Subtract center of gravity from $\vepsilon^{(x)}$ in $\vepsilon = [\vepsilon^{(x)}, \vepsilon^{(h)}]$
    \STATE $\vz_{s} = \frac{1}{\alpha_{t|s}} \vz_t - \frac{\sigma_{t|s}^2}{\alpha_{t|s}\sigma_t} \cdot \phi(\vz_t, t) + \sigma_{t \rightarrow s} \cdot \vepsilon$
\ENDFOR
\STATE Sample $\vx, \vh \sim p(\vx, \vh | \vz_0)$
\end{algorithmic}
\end{algorithm}

\textbf{The Generative Denoising Process}\hspace{4pt}
To define the generative process, the noise posteriors $q(\vz_s | \vx, \vh, \vz_t)$ of Equation~\ref{eq:diffusion_points_features} can be used in the same fashion as in Equation~\ref{eq:background_noise_posterior} by replacing the data variables $\vx, \vh$ by neural network approximations $\hat{\vx}, \hat{\vh}$:
\begin{equation}
    p(\vz_s | \vz_t) = \mathcal{N}_{xh}(\vz_s | \vmu_{t \rightarrow s}([\hat{\vx}, \hat{\vh}], \vz_t), \sigma_{t \rightarrow s}^2 \rmI) \\
    \label{eq:edm_generative}
\end{equation}
where $\hat{\vx}, \hat{\vh}$ depend on $\vz_t, t$ and the neural network $\phi$. As conventional in modern diffusion models, we use the noise parametrization to obtain $\hat{\vx}, \hat{\vh}$. Instead of directly predicting them, the network $\phi$ outputs $\hat{\vepsilon} = [\hat{\vepsilon}^{(x)}, \hat{\vepsilon}^{(h)}]$ which is then used to compute:
\vspace{-7pt}\begin{equation}
    [\hat{\vx}, \hat{\vh}] = \vz_t / \alpha_t - 
      \hat{\vepsilon}_t \cdot \sigma_t / \alpha_t 
    \label{eq:edm_parametrization_epsilon}
\end{equation}
If $\hat{\vepsilon}_t$ is computed by an equivariant function $\phi$ then the denoising distribution in Equation~\ref{eq:edm_generative} is equivariant. To see this, observe that rotating $\vz_t$ to $\rmR\vz_t$ gives $\rmR \hat{\vepsilon}_t = \phi(\rmR \vz_t, t)$. Furthermore, the mean of the denoising equation rotates $\rmR \hat{\vx} = \rmR \vz_t^{(x)} / \alpha_t - \rmR \hat{\vepsilon}_t^{(x)} \sigma_t / \alpha_t$ and since the noise is isotropic, the distribution is equivariant as desired. 

To sample from the model, one first samples $\vz_T \sim \mathcal{N}_{xh}(\mathbf{0}, \rmI)$ and then iteratively samples $\vz_{t-1} \sim p(\vz_{t-1} | \vz_t)$ for $t = T, \ldots, 1$ and then finally samples $\vx, \vh \sim p(\vx, \vh | \vz_0)$, as described in Algorithm~\ref{alg:sample_edm}.

\textbf{Optimization Objective}\hspace{4pt}
Recall that a likelihood term of this model is given by $\mathcal{L}_t = -\mathrm{KL}(q(\vz_s | \vx, \vz_t) || p(\vz_s | \vz_t))$. Analogous to Equation~\ref{eq:loss_t_epsilon_param}, in this parametrization the term simplifies to:
\vspace{-3pt}\begin{equation}
    \mathcal{L}_t = \mathbb{E}_{\vepsilon_t \sim \mathcal{N}_{xh}(0,~ \rmI)} \big{[}\frac{1}{2} ~w(t)~ || \vepsilon_t - \hat{\vepsilon}_t ||^2\big{]},
    \label{eq:loss_t_edm}\vspace{-3pt}
\end{equation}
where $w(t) = (1 - \mathrm{SNR}(t-1) / \mathrm{SNR}(t))$ and $\hat{\vepsilon}_t = \phi(\vz_t, t)$. This is convenient: even though parts of the distribution of $\mathcal{N}_{xh}$ operate on a subspace, the simplification in Equation~\ref{eq:loss_t_epsilon_param} also holds here, and can be computed for all components belonging to $\vx$ and $\vh$ at once. There are three reason why this simplification remains true: firstly, $\mathcal{N}_x$ and $\mathcal{N}$ within $\mathcal{N}_{xh}$ are independent, so the divergence can be separated into two divergences. Further, the KL divergence between the $\mathcal{N}_x$ components are still compatible with the standard KL equation for normal distributions, as they rely on a Euclidean distance (which is rotation invariant) and the distributions are isotropic with equal variance. Finally, because of the similarity in KL equations, the results can be combined again by concatenating the components in $\vx$ and $\vh$. For a more detailed argument see Appendix~\ref{sec:center_gravity_normal}. An overview of the optimization procedure is given in Algorithm~\ref{alg:optimize_edm}.

Following \citep{ho2020denoising} during training we set $w(t) = 1$ as it stabilizes training and it is known to improve sample quality for images. Experimentally we also found this to hold true for molecules: 
even when evaluating the probabilistic variational objective for which $w(t) = (1 - \mathrm{SNR}(t-1) / \mathrm{SNR}(t))$, the model trained with $w(t) = 1$ outperformed models trained with the variational $w(t)$.

In summary, we have defined a diffusion process, a denoising model and an optimization objective between them. To further specify our model, we need to define the neural network $\phi$ that is used within the denoising model.

\subsection{The Dynamics} \label{sec:dynamics}
We learn the E(n) equivariant dynamics function $[\hat{\vepsilon}_t^{(x)}, \hat{\vepsilon}_t^{(h)}] = \phi(\vz_t^{(x)}, \vz_t^{(h)}, t)$ of the diffusion model  using the equivariant network $\mathrm{EGNN}$ introduced in Section \ref{sec:equivariance} in the following way:
$$ \small
   \hat{\vepsilon}_t^{(x)}, \hat{\vepsilon}_t^{(h)} = \mathrm{EGNN} (\vz_t^{(x)}, [\vz_t^{(h)}, t/T]) - [\vz_t^{(x)}, \mathbf{0} \,]
$$
 Notice that we simply input $\vz_t^{(x)}, \vz_t^{(h)}$ to the EGNN with the only difference that $t/T$ is concatenated to the node features. The estimated noise $\hat{\vepsilon}_t^{(x)}$ is given by the output of the EGNN from which the input coordinates $\vz_t^{(x)}$ are removed.
 Importantly, since the outputs have to lie on a zero center of gravity subspace, the component $\hat{\vepsilon}_t^{(x)}$ is projected down by subtracting its center of gravity. This then satisfies the rotational and reflection equivariance on $\hat{\vx}$ with the parametrization in Equation~\ref{eq:edm_parametrization_epsilon}.

\subsection{The Zeroth Likelihood Term}
\label{sec:zeroth_likelihood}
In typical diffusion models \citep{ho2020denoising}, the data being modelled
is ordinal
 which makes the design of $\mathcal{L}_0^{(h)} = \log p(\vh | \vz_0^{(h)})$ relatively simple. Specifically, under very small noise perturbations of the original data distribution $p_{data}(\vh)$ (when $\alpha_0 \approx 1$ and $\sigma_0 \approx 0$) we have
$$q(\vh | \vz_0^{(h)}) = \frac{q(\vz_0^{(h)} | \vh) p_{data}(\vh)}{\sum_{\vh} q(\vz_0^{(h)} | \vh) p_{data}(\vh)} \approx 1,$$ 
when $\vz_0^{(h)}$ is sampled from the noising process $q(\vz_0^{(h)} | \vh)$. Because $q(\vz_0^{(h)} | \vh)$ is a highly peaked distribution, in practice it tends to zero for all but one single discrete state of $\vh$. Furthermore, $p_{data}$ is constant over this small peak, and thus $q(\vh | \vz_0^{(h)}) \approx 1$ when $\vh$ is the closest integer value to $\vz_0^{(h)}$. This can be used to model integer molecular properties such as the atom charge. Following standard practice we let:
\begin{equation}
    p(\vh | \vz_0^{(h)}) = \int_{\vh-\frac{1}{2}}^{\vh+\frac{1}{2}}\mathcal{N}(\vu| \vz_0^{(h)}, \sigma_0) \mathrm{d}\vu,
\end{equation}
which most likely equals $1$ for reasonable noise parameters $\alpha_0$, $\sigma_0$ and it computed as $\Phi((\vh+\frac{1}{2} - \vz_0^{(h)}) / \sigma_0) - \Phi((\vh-\frac{1}{2} - \vz_0^{(h)}) / \sigma_0)$ where $\Phi$ is the CDF of a standard normal distribution. For categorical features such as the atom types, this model would however introduce an undesired bias.

\textbf{Categorical features}\hspace{4pt}
For categorical features such as the atom type, the aforementioned integer representation is unnatural and introduces bias. Instead of using integers for these features, we operate directly on a one-hot representation. Suppose $\vh$ is an array whose values represent categories in $\{c_1, ..., c_d\}$ such as atom types. Then $\vh$ is encoded with a $\operatorname{one-hot}$ function $\vh \mapsto \vh^{\text{onehot}}$ such that
$\vh^\text{onehot}_{i, j} = \mathds{1}_{h_i=c_j}$.
The noising process over $\vz_t^{(h)}$ can then directly be defined using the one-hot representation $\vh^{\text{onehot}}$ equivalent to its definition for integer values, i.e. $q(\vz_t^{(h)} | \vh) = \mathcal{N}(\vz_t^{(h)} | \alpha_t \vh^{\text{onehot}}, \sigma_t^2 \rmI)$ with the only difference that $\vz_t^{(h)}$ has an additional dimension axis with the size equal to the number of categories. Since the data is discrete and the noising process is assumed to be well defined, by the same reasoning as for integer data we can define probability parameters $\rvp$ to be proportional to the normal distribution integrated from $1 - \frac{1}{2}$ to $1 + \frac{1}{2}$. Intuitively, 
when a small amount of noise is sampled and added to the one-hot representation, then the value corresponding to the active class will almost certainly be between $1 - \frac{1}{2}$ and $1 + \frac{1}{2}$: \vspace{-0.3cm}
\begin{equation*}
\small
p(\vh | \vz_0^{(h)}) = \mathcal{C}(\vh | \rvp), \rvp \propto \int_{\boldsymbol{1} - \frac{1}{2}}^{\boldsymbol{1} + \frac{1}{2}}\mathcal{N}(\vu| \vz_0^{(h)}, \sigma_0) \mathrm{d}\vu
\end{equation*}
where $\rvp$ is normalized to sum to one and $\mathcal{C}$ is a categorical distribution. In practice this distribution will almost certainly equal one for values $\vz_0^{(h)}$ that were sampled from the diffusion process given $\vh$.

\textbf{Continuous positions}\hspace{4pt}
For continuous positions, defining $\mathcal{L}_0^{(x)} = \log p(\vx | \vz_0^{(x)})$ is a little more involved than for discrete features. A similar analysis assuming $p_{data}(\vx)$ is constant results in:
$$q(\vx | \vz_0^{(x)}) = \frac{q(\vz_0^{(x)}| \vx)p_{data}(\vx)}{\int_\vx q(\vz_0^{(x)}| \vx)p_{data}(\vx)} \approx \frac{q(\vz_0^{(x)}| \vx)}{\int_\vx q(\vz_0^{(x)}| \vx)},$$
which by completing the square we find is equal to $\mathcal{N}_x(\vx | \vz_0^{(x)} / \alpha_0, \sigma_0^2 / \alpha_0^2 \rmI)$.
Empirically, we find that our model achieves better likelihood performance, especially with lower $\mathrm{SNR}$ rates, when we use $\hat{\vx}$ as a prediction for the mean. After simplifying the terms it turns out that this essentially adds an additional correction term containing the estimated noise $\hat{\vepsilon}_0^{(x)}$, which originates from Equation~\ref{eq:edm_parametrization_epsilon} and can be written as:
\begin{equation}
p(\vx | \vz_0) = \mathcal{N}\Big{(}\vx \big{|} \vz_0^{(x)} / \alpha_0 - \sigma_0 / \alpha_0 \hat{\vepsilon}_0, \sigma_0^2 / \alpha_0^2 \rmI \Big{)}.
\end{equation}
When this parametrization is chosen the log likelihood component $\mathcal{L}_0^{(x)}$ can be re-written to:
\small \begin{equation*}
    \mathcal{L}_0^{(x)} = \mathbb{E}_{\vepsilon^{(x)} \sim \mathcal{N}_x(0, \rmI)} \Big{[} \log Z^{-1} - \frac{1}{2}||\vepsilon^{(x)} - \phi^{(x)}(\vz_0, 0) ||^2 \Big{]},
\end{equation*} \normalsize
with the normalization constant $Z$. Conveniently, this allows Equation~\ref{eq:loss_t_edm} for losses $\mathcal{L}_t$ to also be used for $t=0$ in its $\vx$ component, by defining $w(0) = -1$. The normalization constant then has to be added separately. This normalization constant $Z = (\sqrt{2 \pi} \cdot \sigma_0 / \alpha_0)^{(M-1)\cdot n}$ where the $(M-1)\cdot n$ arises from the zero center of gravity subspace is described in Appendix~\ref{sec:center_gravity_normal}.

\textbf{Scaling Features}\hspace{4pt}
Since coordinates, atom types and charges represent different physical quantities, we can define a relative scaling between them. While normalizing the features simply makes them easier to process for the neural network, the relative scaling has a deeper impact on the model: when the features $\vh$ are defined on a smaller scale than the coordinates $\vx$, the denoising process tends to first determine rough positions and decide on the atom types only afterwards. Whereas scaling $\vx$ requires a correction in the log-likelihood since it is continuous, scaling $\vh$ does not require a correction and is not problematic as long as the difference in discrete values is large compared to $\sigma_0$. We find empirically that
defining the input to our EDM model as $[\vx, 0.25 ~\vh^\text{onehot}, 0.1~ \vh^\text{atom charge}]$ significantly improves performance over non-scaled inputs. 

\textbf{Number of Atoms}\hspace{4pt}
In the above sections we have considered the number of atoms $M$ to be known beforehand. To adapt to different molecules with different sizes, we compute the categorical distribution $p(M)$ of molecule sizes on the training set. To sample from the model $p(\vx, \vh, M)$, $M \sim p(M)$ is first sampled and then $\vx, \vh \sim p(\vx, \vh | M)$ are sampled from the EDM. For clarity this conditioning on $M$ is often omitted, but it remains an important part of the generative process and likelihood computation.

\subsection{Conditional generation} \label{sec:conditional_generation}
In this section we describe a straightforward extension to the proposed method to do conditional generation $\vx, \vh \sim p(\vx, \vh | c)$ given some desired property $c$. We can define the optimization lower bound for the conditional case as
$
    \log p(\vx, \vh | c) \geq \mathcal{L}_{c,0} + \mathcal{L}_{c,\text{base}} + \sum_{t=1}^{T} \mathcal{L}_{c,t}
$, where the different $\mathcal{L}_{c,t}$ for $1 \leq t < T-1$ are defined similarly to Equation~\ref{eq:loss_t_edm}, 
with the important difference that the function $\hat{\vepsilon}_t = \phi(\vz_t, [t,c])$ takes as additional input a property $c$ which is concatenated to the nodes features. Given a trained conditional model we define the generative process by first sampling the number of nodes $M$ and a property value $c$ from a parametrized distribution $c, M \sim p(c, M)$ defined in Appendix~\ref{ap:conditional_exp}. Next, we can generate molecules $\vx, \vh$ given $c$, $M$ using our Conditional EDM $\vx, \vh \sim p(\vx, \vh | c, M)$. 

\section{Related Work}

Diffusion models \citep{sohldickstein2015diffusion} are generative models that have recently been connected to score-based methods via denoising diffusion models \citep{song2019estimatinggradients,ho2020denoising}. This new family of generative models has proven to be very effective for the generation of data such as images \citep{ho2020denoising,nichol2021improved}. 

Some recent methods directly generate molecules in 3D: \citep{gebauer2019symmetry, luo2021autoregressive, luo20213d, gebauer2021inverse} define an order-dependent autoregressive distribution from which atoms are iteratively sampled. \citep{ragoza2020learning} maps atoms to a fixed grid and trains a VAE using 3D convolutions. E-NF \citep{satorras2021en_flows} defines an equivariant normalizing flow that integrates a differential equation. Instead, our method learns to denoise a diffusion process, which scales better during training.

A related branch of literature is concerned by solely predicting coordinates from molecular graphs, referred to as the conformation. Examples of such methods utilize conditional VAEs \citep{simm2019generative}, Wasserstein GANs \citep{hoffmann2019generating}, and normalizing flows \citep{noe2019boltzmann}, with adaptions for Euclidean symmetries in \citep{kohler2020equivariantflows, xu2021end, simm2021symmetryaware, ganea2021geomol, guan2022energyinspired} resulting in performance improvements. In recent works \citep{shi2021learninggradientfieldsmolecular,luo2021predictingmolecularscore,xu2022geodiff} it was shown that score-based and diffusion models are effective at coordinate prediction, especially when the underlying neural network respects the symmetries of the data. Our work can be seen as an extension of these methods that incorporates discrete atom features, and furthermore derives the equations required for log-likelihood computation. In the context of diffusion for discrete variables, unrelated to molecule modelling, discrete diffusion processes have been proposed \citep{sohldickstein2015diffusion,hoogeboom2021argmax,austin2021structured}. However, for 3D molecule generation these would require a separate diffusion process for the discrete features and the continuous coordinates. Instead we define a joint process for both of them.

 Tangentially related, other methods generate molecules in graph representation. Some examples are autoregressive methods such as \citep{liu2018constrained, you2018graph, liao2019efficient}, and one-shot approaches such as \citep{simonovsky2018graphvae, de2018molgan,bresson2019two,kosiorek2020conditional, krawczuk2021gggan}. However such methods do not provide conformer information which is useful for many downstream tasks.

\begin{figure*}[t!]
\centering
\includegraphics[width=.99\textwidth]{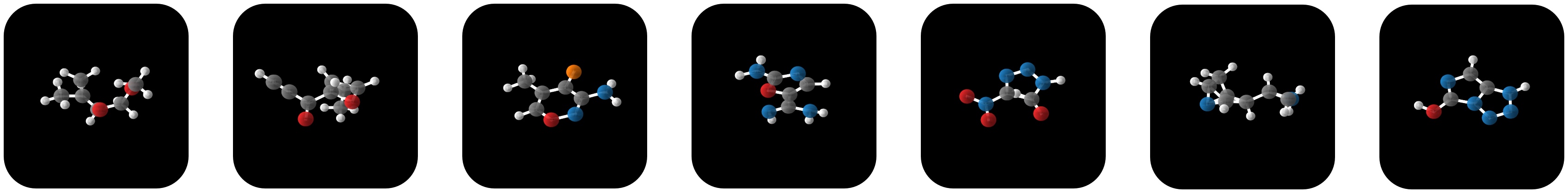}
\includegraphics[width=1.0\textwidth]{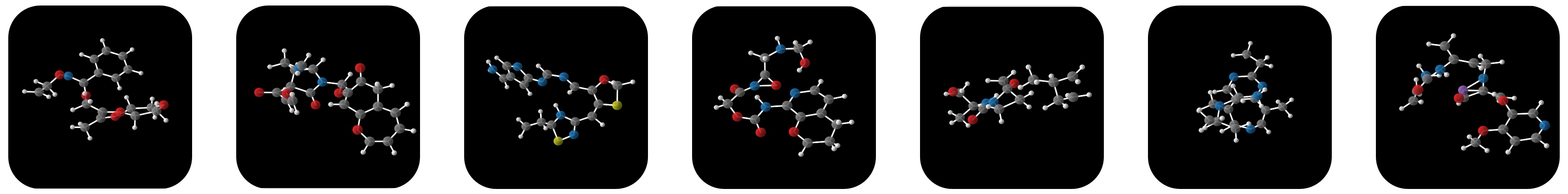}

\vspace{-10pt}
\caption{Selection of samples generated by the denoising process of our EDM trained on QM9 (up) and GEOM-DRUGS (down). 
}
\label{fig:molecule_samples_qm9+geom}
\vspace{-7pt}
\end{figure*}

\section{Experiments}

\subsection{Molecule Generation | QM9}

QM9 \citep{ramakrishnan2014quantum} is a standard dataset that contains molecular properties and atom coordinates for 130k small molecules with up to 9 heavy atoms (29 atoms including hydrogens). In this experiment we train EDM to unconditionally generate molecules with 3-dimensional coordinates, atom types (H, C, N, O, F) and integer-valued atom charges. We use the train/val/test partitions introduced in \citep{NEURIPS2019_03573b32}, which consists of 100K/18K/13K samples respectively for each partition. 

\textbf{Metrics}
Following \citep{satorras2021en_flows}, we use the distance between pairs of atoms and the atom types to predict bond types (single, double, triple or none).
We then measure atom stability (the proportion of atoms that have the right valency) and molecule stability (the proportion of generated molecules for which all atoms are stable).

\textbf{Baselines:} We compare EDM to two existing E(3) equivariant models: G-Schnet \citep{gebauer2019symmetry} and Equivariant Normalizing Flows (E-NF) \citep{satorras2021en_flows}. For G-Schnet we extracted $10000$ samples from the publicly available code to run the analysis. In order to demonstrate the benefits of equivariance, we also perform an ablation study and run a non-equivariant variation of our method that we call Graph Diffusion Models (GDM). The Graph diffusion model is run with the same configuration as our method, except that the EGNN is replaced by a non-equivariant graph network defined in Appendix \ref{appendix:exp}. We also experiment with GDM-aug, where the GDM model is trained on data augmented with random rotations. All models use 9 layers, 256 features per layer and SiLU activations. They are trained using Adam with batch size 64 and learning rate $10^{-4}$.

\textbf{Results} are reported in Table \ref{tab:qm9_results}. Our method outperforms previous methods (E-NF and G-Schnet), as well as its non-equivariant counterparts on all metrics. It is interesting to note that the negative log-likelihood of the EDM is much lower than other models, which indicates that it is able to create sharper peaks in the model distribution.

 \begin{table}[t]
     \vspace{-8pt}
    \centering
    \caption{Neg. log-likelihood $-\log p(\rvx, \rvh, M)$, atom stability and molecule stability with standard deviations across 3 runs on QM9, each drawing 10000 samples from the model.}\label{tab:qm9_results}
        \scalebox{.9}{
    \begin{tabular}{l r r r}
    \toprule
     \# Metrics & NLL\hspace{0.58cm} & Atom stable (\%) & Mol stable (\%)\\
      \midrule
        {E-NF} & -59.7\hspace{0.58cm} & 85.0\hspace{0.58cm} & 4.9\hspace{0.58cm} \\
      \vspace{0.1cm}  {G-Schnet} & N.A\hspace{0.65cm} & 95.7\hspace{0.58cm} & 68.1\hspace{0.58cm}  \\
        GDM & -94.7\hspace{0.58cm} & 97.0\hspace{0.58cm} & 63.2\hspace{0.58cm}  \\ 
        GDM-aug & -92.5\hspace{0.58cm} & 97.6\hspace{0.58cm} & 71.6\hspace{0.58cm} \\ 
        EDM (ours) & \textbf{-110.7}$\spm{1.5}$ & \textbf{98.7}$\spm{0.1}$ & \textbf{82.0}$\spm{0.4}$ \\ 
        \midrule
        Data &  & 99.0\hspace{0.58cm} & 95.2\hspace{0.58cm} \\
     \bottomrule
    \end{tabular}}
     \vspace{-15pt}
  \end{table}

 \begin{table}[]
    \centering
    \vspace{-8pt}
    \caption{Validity and uniqueness over 10000 molecules with standard deviation across 3 runs. Results marked (*) are not directly comparable, as they do not use 3D coordinates to derive bonds. \\ \small{H: model hydrogens explicitly}}
    \scalebox{.9}{
    \begin{tabular}{l c r r}
    \toprule
    Method    & H  & Valid (\%) & Valid and Unique (\%)\\ \midrule
    Graph VAE (*) &    &   $55.7$\hspace{0.58cm} & $42.3$\hspace{0.58cm} \\
    GTVAE (*)     &      &  $74.6$\hspace{0.58cm} & $16.8$\hspace{0.58cm}   \\
    Set2GraphVAE (*)  &  & $59.9 \spm{1.7}$ & $56.2 \spm{1.4}$  \\ 
    \vspace{0.1cm}
    EDM (ours) &  & $\bm{97.5} \spm{0.2}$ & $\bm{94.3} \spm{0.2}$  \\
    E-NF & \checkmark & $40.2$\hspace{0.58cm} & $39.4$\hspace{0.58cm} \\
    G-Schnet & \checkmark  & $85.5$\hspace{0.58cm}    & $80.3$\hspace{0.58cm} \\
    GDM-aug & \checkmark & $90.4$\hspace{0.58cm} & $89.5$\hspace{0.58cm} \\
    EDM (ours) & \checkmark & $\bm{91.9} \spm{0.5}$ &  $\bm{90.7} \spm{0.6}$  \\ 
    \midrule 
    Data & \checkmark & $97.7$\hspace{0.58cm} & $97.7$\hspace{0.58cm}  \\ \bottomrule
    \end{tabular}}
    \label{tab:qm9}
    \vspace{-8pt}
\end{table}

Further, EDMs are compared to one-shot graph-based molecule generation models that do not operate on 3D coordinates: GraphVAE \citep{simonovsky2018graphvae}, GraphTransformerVAE \citep{mitton2021graph}, and Set2GraphVAE \citep{vignac2021top}. For G-Schnet and EDM, the bonds are directly derived from the distance between atoms. We report validity (as measured by RDKit) and uniqueness of the generated compounds. Following \citep{vignac2021top} novelty is not included here. For a discussion on the issues with the novelty metric, see Appendix~\ref{appendix:exp}. As can be seen in Table~\ref{tab:qm9}, the EDM is able to generate a very high rate of valid and unique molecules. This is impressive since the 3D models are at a disadvantage in this metric, as the rules to derive bonds are very strict. Interestingly, even when including hydrogen atoms in the model, the performance of the EDM does not deteriorate much. A possible explanation is that the equivariant diffusion model scales effectively and learn very precise distributions, as evidenced by the low negative log-likelihood.  


\subsection{Conditional Molecule Generation}

\begin{figure*}[t!]
\includegraphics[width=1.0\textwidth]{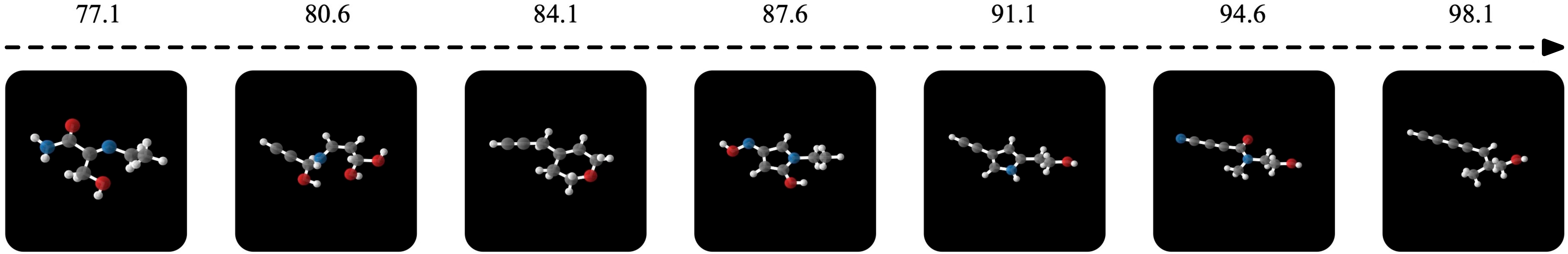}
\centering
\vspace{-17pt}
\caption{Generated molecules by our Conditional EDM when interpolating among different Polarizability $\alpha$ values with the same reparametrization noise $\vepsilon$. Each $\alpha$ value is provided on top of each image.
}
\label{fig:conditional_main}
\vspace{-13.pt}
\end{figure*}

\begin{table}[t] 
\vspace{-5pt}
\setlength{\tabcolsep}{2.5pt}
\footnotesize
    \centering
    \caption{Mean Absolute Error for molecular property prediction by a EGNN classifier $\phi_c$ on a QM9 subset, EDM generated samples and two different baselines "Naive (U-bounds)" and "\# Atoms".}
    \begin{tabular}{l r r r r r r}
    \toprule
    Task & $\alpha$& $\Delta \varepsilon$ & $\varepsilon_{\mathrm{HOMO}}$ & $\varepsilon_{\mathrm{LUMO}}$ & $\mu$ & $C_v$\\
    Units & Bohr$^3$ & meV & meV & meV & D & $\frac{\text{cal}}{\text{mol}}$K  \\
    \midrule
    Naive (U-bound) & 9.01  &  1470 & 645 & 1457  & 1.616  & 6.857   \\
    \#Atoms         & 3.86  & 866 & 426 & 813 & 1.053  &  1.971 \\
    EDM             & 2.76  & 655 & 356 & 584 & 1.111 & 1.101 \\
    QM9 (L-bound)   & 0.10  & 64 & 39 & 36 & 0.043 & 0.040  \\
    \bottomrule
    \end{tabular}
    \label{tab:conditional_results}
    \vspace{-0.6cm}
\end{table}

In this section, we aim to generate molecules targeting some desired properties. This can be of interest towards the process of drug discovery where we need to obtain molecules that satisfy specific properties. We train our conditional diffusion model from Section \ref{sec:conditional_generation} in QM9 conditioning the generation on properties $\alpha$, gap, homo, lumo, $\mu$ and $C_v$ described in more detail in Appendix \ref{ap:conditional_exp}. In order to assess the quality of the generated molecules w.r.t. to their conditioned property, we use the property classifier network $\phi_c$ from \citet{satorras2021egnn}. We split the QM9 training partition into two halves ${D_a, D_b}$ of 50K samples each. The classifier $\phi_c$ is trained on the first half $D_a$, while the Conditional EDM is trained on the second half $D_b$. Then, $\phi_c$ is evaluated on the EDM conditionally generated samples. We also report the loss of $\phi_c$ on $D_b$ as a lower bound named "QM9 (L-bound)". The better EDM approximates $D_b$ the smaller the gap between "EDM" and "QM9 (L-bound)". Further implementation details are reported in Appendix \ref{ap:conditional_exp}.

\textbf{Baselines:} We provide two baselines in which molecules are to some extent agnostic to their respective property $c$. In the first baseline we simply remove any relation between molecule and property by shuffling the property labels in $D_b$ and then evaluating $\phi_c$ on it. We name this setting "Naive (Upper-Bound)". The second baseline named "\#Atoms" predicts the molecular properties in $D_b$ by only using the number of atoms in the molecule. If "EDM" overcomes "Naive (Upper-Bound)" it should be able to incorporate conditional property information into the generated molecules. If it overcomes "\#Atoms" it should be able to incorporate it into the molecular structure beyond the number of atoms.

\textbf{Results (quantitative):} Results are reported in Table \ref{tab:conditional_results}. EDM outperforms both "Naive (U-bound)" and "\#Atoms" baselines in all properties (except $\mu$) indicating that it is able to incorporate property information into the generated molecules beyond the number of atoms for most properties. However, we can see there is still room for improvement by looking at the gap between "EDM" and "QM9 (L-bound)".

\textbf{Results (qualitative):} In Figure \ref{fig:conditional_main}, we interpolate the conditional generation among different Polarizability values $\alpha$ while keeping the noise $\vepsilon$ fixed. The Polarizability is the tendency of a molecule to acquire an electric dipole moment when subject to an external electric field. We can expect less isometrically shaped molecules for large $\alpha$ values. This is the obtained behavior in Figure \ref{fig:conditional_main} -- we show that this behavior is consistent across different runs in Appendix \ref{ap:conditional_exp}.

\subsection{GEOM-Drugs}
 \begin{table}[t]
 \vspace{-0.25cm}
    \centering
    \caption{Neg. log-likelihood, atom stability and Wasserstein distance between generated and training set energy distributions.}\label{tab:geom_results}
    \begin{tabular}{ l r r r }
    \toprule
     \# Metrics & NLL & Atom stability (\%)& $\mathcal W$ \\
      \midrule
         GDM & $-~14.2$ & $75.0$ & $3.32$ \\ 
         GDM-aug & $-~58.3$ & $77.7$ & $4.26$\\ 
         \vspace{0.1 cm} EDM & $\bm{-137.1}$ & $\bm{81.3}$ &  $\bm{1.41}$  \\ 
        Data & & $86.5$ & $0.0$\\
     \bottomrule
    \end{tabular}
    \vspace{-.6cm}
  \end{table}

While QM9 features only small molecules, GEOM \citep{axelrod2020geom} is a larger scale dataset of molecular conformers. It features 430,000 molecules with up to 181 atoms and 44.4 atoms on average. For each molecule, many conformers are given along with their energy. From this dataset we retain the 30 lowest energy conformations for each molecule. The models learn to generate the 3D positions and atom types of these molecules. All models use 4 layers, 256 features per layer, and are trained using Adam with batch size 64 and learning rate $10^{-4}$.

Since molecules in this dataset are bigger and have more complex structures, predicting the bond types using the atom types and the distance between atoms with lookup tables results in more errors than on QM9. For this reason, we only report the atom stability, which measures $86.5\%$ stable atoms on the dataset. Intuitively, this metric describes the percentage of atoms that have bonds in typical ranges -- ideally, generative models should generate a comparable number of stable atoms. We also measure the energy of generated compounds with the software used to generate the conformations of the GEOM dataset \citep{bannwarth2019gfn2}. After computing the energies of sampled molecules and the dataset, we measure the Wasserstein distance between their histograms. In Table~\ref{tab:geom_results} we can see that the EDM outperforms its non-equivariant counterparts on all metrics. In particular, EDM is able to capture the energy distribution well, as can be seen on the histograms in Appendix~\ref{appendix:exp}.

\section{Conclusions}

We presented EDM, an E(3) equivariant diffusion model for molecule generation in 3D. While previous non-autoregressive models mostly focused on very small molecules with up to 9 atoms, our model scales better and can generate valid conformations while explicitly modeling hydrogen atoms. We also evaluate our model on the larger GEOM-DRUGS dataset, setting the stage for models for drug-size molecule generation in 3D.

\bibliography{example_paper}
\bibliographystyle{icml2021}

\newpage
\clearpage
\onecolumn

\appendix

\section{The zero center of gravity, normal distribution}
\label{sec:center_gravity_normal}

Consider the Euclidean variable $\vx \in \mathbb{R}^{M \times n}$ in the linear subspace $\sum_i \vx_i = \mathbf{0}$. In other words, $\vx$ is a point cloud where its center of gravity is zero. One can place a normal distribution $\mathcal{N}_x$ over this subspace and its likelihood can be expressed as:
\begin{equation*}
    \mathcal{N}_x(\vx | \vmu, \sigma^2 \rmI) = (\sqrt{2 \pi} \sigma)^{-(M-1)\cdot n} \exp \Big{(} -\frac{1}{2\sigma^2} ||\vx - \vmu||^2 \Big{)}
\end{equation*}
Here $\vmu$ also lies in the same subspace as $\vx$. Also note a slight abuse of notation: $\vx, \vmu$ are technically two-dimensional matrices but are treated in the distribution as single-dimensional (flattened) vectors. To sample from this distribution, there are multiple options. For instance, one could sample from a normal distribution with dimensionality $(M-1)\cdot n$ and then map the sample to the $M\cdot n$ dimensional ambient space so that its center of gravity equals zero. However there is an easier alternative: One can sample in the $M\cdot n$ dimensional ambient space directly, and subtract $\sum_i \vx_i$. Because the normal distributions are isotropic (meaning its variance in any direction you pick is $\sigma^2$) this is equivalent to the aforementioned method. More detailed analysis are given in \citep{satorras2021en_flows} and \citep{xu2022geodiff}.

\paragraph{KL Divergence}
A standard KL divergence for between two isotropic normal distributions $q = \mathcal{N}(\vmu_1, \sigma_1^2 \rmI)$ and $p = \mathcal{N}(\vmu_2, \sigma_2^2 \rmI)$ is given by:
\begin{equation}
    \mathrm{KL}(q || p) = d \cdot \log \frac{\sigma_2}{\sigma_1} + \frac{1}{2} \Big{[} \frac{d \cdot \sigma_1^2 + ||\vmu_1 - \vmu_2||^2}{\sigma_2^2} - d  \Big{]},
    \label{eq:kl_isotropic_normals}
\end{equation}
where $d$ is the dimensionality of the distribution. Recall that in our case the diffusion and denoising process have the same variance $\sigma_{Q,s,t}^2$. If $\sigma_1 = \sigma_2 = \sigma$, then the KL divergence simplifies to:
\begin{equation}
    \mathrm{KL}(q || p) = \frac{1}{2} \Big{[} \frac{||\vmu_1 - \vmu_2||^2}{\sigma^2} \Big{]}.
    \label{eq:kl_same_stdev}
\end{equation}
Suppose now that $\mathcal{N}_1(\tilde{\vmu}_1, \sigma \rmI)$ and $\mathcal{N}_2(\tilde{\vmu}_2, \sigma \rmI)$ are defined on a linear subspace, where the mean $\tilde{\vmu}$ is defined with respect to any coordinate system in the subspace. The KL divergence between these distributions then includes a term containing the Euclidean distance $||\tilde{\vmu}_1 - \tilde{\vmu}_2||^2$

Similar to the arguments in \citep{satorras2021en_flows,xu2022geodiff}, an orthogonal transformation $\rmQ$ can be constructed that maps an ambient space where $\sum_i \vmu_i = \mathbf{0}$ to the subspace in such a way that $\begin{bmatrix} \tilde{\vmu} \\ \mathbf{0} \end{bmatrix} = \rmQ \vmu$. Observe that $||\tilde{\vmu}|| = ||\begin{bmatrix} \tilde{\vmu} \\ \mathbf{0} \end{bmatrix}|| = ||\vmu||$, and therefore $||\tilde{\vmu}_1 - \tilde{\vmu}_2||^2 = ||\vmu_1 - \vmu_2||^2$. This shows that Equation~\ref{eq:kl_same_stdev} can be consistently computed in the ambient space. This also shows an important caveat: in some diffusion models, different variances are used in the posterior of the diffusion process and the denoising process. In those cases one can see from Equation~\ref{eq:kl_isotropic_normals} that the divergence depends on the dimensionality of the subspace, not to be confused with the dimensionality of the ambient space. 

\paragraph{The combined KL divergence for positions and features}
In the previous section we have shown that the KL divergence for distributions such as $\mathcal{N}_x$, can still be computed in the ambient space as long as standard deviations between two such distributions are the same. Let us know consider the combined KL divergence for distributions $q = \mathcal{N}_{xh}(\vmu_1, \sigma^2 \rmI)$ and $p = \mathcal{N}_{xh}(\vmu_2, \sigma^2 \rmI)$. Note that here the means consist of two parts $\vmu = [\vmu^{(x)}, \vmu^{(h)}]$ where the $x$ part lies in a subspace and the $h$ part is defined freely. The distributions factorize as $\mathcal{N}_{xh}(\vmu, \sigma^2 \rmI) = \mathcal{N}_{x}(\vmu^{(x)}, \sigma^2 \rmI) \cdot \mathcal{N}(\vmu^{(h)}, \sigma^2 \rmI)$. Then the KL divergence simplifies as:
\begin{align}
\begin{split}
    \mathrm{KL}(q||p) &= \mathrm{KL}\Big{(}\mathcal{N}_{x}(\vmu_1^{(x)}, \sigma^2 \rmI) || \mathcal{N}_{x}(\vmu_2^{(x)}, \sigma^2 \rmI  )\Big{)} + \mathrm{KL}\Big{(}\mathcal{N}(\vmu_1^{(h)}, \sigma^2 \rmI) || \mathcal{N}(\vmu_2^{(h)}, \sigma^2 \rmI  )\Big{)} \\
    &= \frac{1}{2} \Big{[} \frac{||\vmu_1^{(x)} - \vmu_2^{(x)}||^2}{\sigma^2} \Big{]} + \frac{1}{2} \Big{[} \frac{||\vmu_1^{(h)} - \vmu_2^{(h)}||^2}{\sigma^2} \Big{]} = \frac{1}{2} \Big{[} \frac{||\vmu_1 - \vmu_2||^2}{\sigma^2} \Big{]}.
\end{split}
\end{align}
Here we have used that products of independent distributions sum in their independent KL terms, and that the sum of the Euclidean distance of two vectors squared is equal to the squared Euclidean distance of the two vectors concatenated. In summary, even though parts of our distribution are defined on a linear subspace, all computation for the KL divergences is still consistent and does not require special treatment. This is however only valid under the condition that the variances of the denoising process and posterior noising process are the same.

\section{Additional Details for the Method}

\textbf{Noise schedule:}
A diffusion process requires a definition for $\alpha_t, \sigma_t$ for $t = 0, \ldots, T$. Since $\alpha_t = \sqrt{1 - \sigma_t^2}$, it suffices to define $\alpha_t$. The values should monotonically decrease, starting $\alpha_0 \approx 1$ and ending at $\alpha_T \approx 0$. In this paper we let
$$\alpha_t = (1 - 2s) \cdot f(t) + s \text{ where } f(t) = (1 - (t/T)^2),$$ for a precision value $10^{-5}$ that avoids numerically unstable situations. This schedule is very similar to the cosine noise schedule introduced in \citep{nichol2021improved}, but ours is somewhat simpler in notation. To avoid numerical instabilities during sampling, we follow the clipping procedure of \citep{nichol2021improved} and compute $\alpha_{t|t-1} = \alpha_t / \alpha_{t-1}$, where we define $\alpha_{-1} = 1$. The values $\alpha_{t|t-1}^2$ are then clipped from below by $0.001$. This avoids numerical instability as $1 / \alpha_{t|t-1}$ is now bounded during sampling. Then the $\alpha_t$ values can be recomputed using the cumulative product $\alpha_t = \prod_{\tau=0}^t \alpha_{\tau|\tau-1}$.

Recall that $\mathrm{SNR}(t) = \alpha_t^2 / \sigma_t^2$. As in \citep{kingma2021variational}, we compute the negative log $\mathrm{SNR}$ curve defined as $\gamma(t) = -(\log \alpha_t^2 - \log \sigma_t^2)$ for $\sigma_t^2 = 1 - \alpha_t^2$. $\gamma(t)$ is a monotonically increasing function from which all required components can be computed with high numerical precision. For instance, $\alpha_t^2 = \mathrm{sigmoid}(-\gamma(t))$, $\sigma_t^2 = \mathrm{sigmoid}(\gamma(t))$, and $\mathrm{SNR}(t) = \exp(-\gamma(t))$.

\textbf{Log-likelihood estimator:}
As discussed, the simplified objective described in Algorithm~\ref{alg:optimize_edm} is optimized during training. However, when evaluating the log-likelihood of samples, the true weighting $w(t) = 1 - \mathrm{SNR}(t-1) / \mathrm{SNR}(t)$ needs to be used. For this purpose, we follow the procedure described in Algorithm~\ref{alg:likelihood_edm}. An important detail is that we choose to put an estimator over $\mathcal{L}_t$ for $t =1, \ldots, T$ using $\mathbb{E}_{t \sim \mathcal{U}(1, \ldots, T)}[T \cdot \mathcal{L}_t] = \sum_{t=1}^T \mathcal{L}_t$, but we require an additional forward pass for $\mathcal{L}_0$. In initial experiments, we found the contribution of $\mathcal{L}_0$ very large compared to other loss terms, which would result in very high variance of the estimator. For that reason, the $\mathcal{L}_0$ is always computed at the expense of an additional forward pass. The resulting $\hat{\mathcal{L}}$ is an unbiased estimator for the log-likelihood.

\begin{algorithm}[H]
   \caption{Log-likelihood estimator for EDMs}
   \label{alg:likelihood_edm}
\begin{algorithmic}
\STATE {\bfseries Input:} Data point $\vx$, neural network $\phi$
\STATE Sample $t \sim \mathcal{U}(1, \ldots, T)$, $\vepsilon_t \sim \mathcal{N}(\mathbf{0}, \rmI)$, subtract center of gravity from $\vepsilon^{(x)}_t$ in $\vepsilon_t = [\vepsilon^{(x)}_t, ~\vepsilon^{(h)}_t$]
\STATE $\vz_t = \alpha_t [\vx, \vh] + \sigma_t \vepsilon_t$
\STATE $\mathcal{L}_t = \frac{1}{2}(1 - \mathrm{SNR}(t-1) / \mathrm{SNR}(t)) ||\vepsilon_t - \phi(\vz_t, t)||^2$
\STATE Sample $\vepsilon_0 \sim \mathcal{N}(\mathbf{0}, \rmI)$, subtract center of gravity from $\vepsilon^{(x)}_0$ in $\vepsilon_0 = [\vepsilon^{(x)}_0, \vepsilon^{(h)}_0]$
\STATE $\vz_0 = \alpha_0 [\vx, \vh] + \sigma_0 \vepsilon_0$
\STATE $\mathcal{L}_0 = \mathcal{L}_0^{(x)} + \mathcal{L}_0^{(h)} = -\frac{1}{2} ||\vepsilon - \phi(\vz_0, 0)||^2 - \log Z + \log p(\vh | \vz_0^{(h)})$
\STATE $\mathcal{L}_{\text{base}} = -\mathrm{KL}(q(\vz_T | \vx, \vh) | p(\vz_T)) = -\mathrm{KL}(\mathcal{N}_{xh}(\alpha_T [\vx, \vh], \sigma_T^2 \rmI) | \mathcal{N}_{xh}(\mathbf{0}, \rmI))$
\STATE Return $\hat{\mathcal{L}} = T \cdot \mathcal{L}_t + \mathcal{L}_0 + \mathcal{L}_{\text{base}}$
\end{algorithmic}
\end{algorithm}

\paragraph{The Dynamics} \label{ap:the_dynamics}

In Section \ref{sec:dynamics} we explained that the dynamics of our proposed Equivariant Diffusion Model (EDM) are learned by the EGNN introduced in Section \ref{sec:equivariance}. The EGNN consists of a sequence of Equivariant Graph Convolutional Layers (EGCL). The EGCL is defined in Eq. \ref{eq:coord_update}. All its learnable components $\phi_e$, $\phi_h$, $\phi_x$, $\phi_{inf}$ by Multilayer Perceptrons:

\textbf{Edge operation} $\phi_e$. Takes as input two node embeddings. The squared distance $d_{ij}^2 = \| \vx_{i}^{l}-\vx_{j}^{l} \|_2^2$, and the squared distance at the first layer as the optional attribute $a_{ij} = \| \vx_{i}^{0}-\vx_{j}^{0} \|_2^2$ and outputs $\rmm_{ij} \in \mathbb{R}^{\text{nf}}$.

$\qquad \mathrm{concat}[\vh_{i}^{l}, \vh_{j}^{l}, d_{ij}^2, a_{ij}] \xrightarrow{} \{\mathrm{Linear}(\text{nf}\cdot 2 + 2, \text{nf}) \xrightarrow{} \mathrm{Silu} \xrightarrow{} \mathrm{Linear}(\text{nf}, \text{nf}) \xrightarrow{} \mathrm{Silu} \} \xrightarrow{} \rmm_{ij}$ 

\textbf{Edge inference operation} $\phi_{inf}$. Takes as input the message $\rmm_{ij}$ and outputs a scalar value $\tilde{e}_{ij} \in (0, 1)$.
 
 $\qquad \rmm_{ij} \xrightarrow{}  \{\mathrm{Linear}(\text{nf}, 1) \xrightarrow{} \text{Sigmoid}\} \xrightarrow{} \tilde{e}_{ij}$

 \textbf{Node update $\phi_h$} Takes as input a node embedding and the aggregated messages and outputs the updated node embedding.
 
 $\qquad \mathrm{concat}[\vh_{i}^{l}, \rmm_{ij}] \xrightarrow{} \{\mathrm{Linear}(\text{nf}\cdot 2, \text{nf}) \xrightarrow{} \text{Silu} \xrightarrow{} \mathrm{Linear}(\text{nf}, \text{nf}) \xrightarrow{} \mathrm{add}(\cdot, \vh_{i}^{l})\} \xrightarrow{} \vh_{i}^{l+1}$
 
 \textbf{Coordinate update} $\phi_x$. Has the same inputs as $\phi_e$ and outptus a scalar value.
 
 $\qquad \mathrm{concat}[\vh_{i}^{l}, \vh_{j}^{l}, d_{ij}^2, a_{ij}] \xrightarrow{} \{\mathrm{Linear}(\text{nf}\cdot 2 + 2, \text{nf}) \xrightarrow{} \mathrm{Silu} \xrightarrow{} \mathrm{Linear}(\text{nf}, \text{nf}) \xrightarrow{} \mathrm{Silu} \xrightarrow{} \mathrm{Linear}(\text{nf}, \text{1}) \} \xrightarrow{} \text{Output}$

\paragraph{Equivariant Processes}
To be self-contained, a version of the proof from \citep{xu2022geodiff} is given here. It shows that if the transition distributions $p(\vz_{t-1} | \vz_t)$ are equivariant and $p(\vz_T)$ is invariant, then every marginal distribution $p(\vz_t)$ is invariant which importantly includes $p(\vz_0)$. Here induction is used to derive the result.

Base case: Observe that $p(\vz_T) = \mathcal{N}(\mathbf{0}, \rmI)$ is equivariant with respect to rotations and reflections, so $p(\vz_T) = p(\rmR \vz_T)$.

Induction step: For some $t \in \{1, \ldots, T\}$ assume $p(\vz_t)$ to be invariant meaning that $p(\vz_t) = p(\rmR \vz_t)$ for all orthogonal $\rmR$. Let $p(\vz_{t-1} | \vz_t)$ be equivariant meaning that $p(\vz_{t-1} | \vz_t) = p(\rmR \vz_{t-1} | \rmR \vz_t)$ for orthogonal $\rmR$. Then:
\begin{align*}
    p(\rmR \vz_{t-1}) &= \int_{\vz_t} p(\rmR \vz_{t-1} | \vz_t) p(\vz_t)  &\text{ Probability Chain Rule} \\ 
    &= \int_{\vz_t} p(\rmR \vz_{t-1} | \rmR \rmR^{-1} \vz_t) p(\rmR \rmR^{-1} \vz_t)  &\text{ Multiply by $\rmR \rmR^{-1} = \rmI$ } \\ 
    &= \int_{\vz_t} p( \vz_{t-1} | \rmR^{-1} \vz_t) p( \rmR^{-1} \vz_t) &\text{ Equivariance \& Invariance } \\ 
    &= \int_{\vu} p( \vz_{t-1} | \vu) p( \vu) \cdot \underbrace{\det \rmR}_{=1} &\text{ Change of Variables $\vu = \rmR^{-1} \vz_t$} \\ &= p(\vz_{t-1}),
\end{align*}
and thus $p(\vz_{t-1})$ is invariant. By induction, $p(\vz_{T-1}), \ldots, p(\vz_{0})$ are all invariant. Compared to \citep{xu2022geodiff}, this proof makes explicit the dependency on a change of variables to rotate the reference frame of integration.

\newpage
\section{Additional Details on Experiments} \label{appendix:exp}

\paragraph{Baseline model}
While our EDM model is parametrized by an $E(3)$ equivariant EGNN network, the GDM model used for the ablation study uses a non equivariant graph network. In this network, the coordinates are simply concatenated with the other node features: $\tilde \vh^0_i = [\vx_i, \vh]$. A message passing neural network \citep{gilmer2017neural} is then applied, that can be written:
\begin{equation*}
\tilde \vh_{i}^{l+1} = \phi_{h}(\tilde \vh_{i}^l, { \sum_{j \neq i}} \tilde{e}_{ij} \rmm_{ij}) \qquad \text{for} \quad \rmm_{ij} = \phi_{e}\left(\tilde \vh_{i}^{l}, \tilde \vh_{j}^{l}, a_{ij}\right) \\
\end{equation*}
The MLPs $\phi_e$, $\phi_h$ are parametrized in the same way as in EGNN, with the sole exception that the input dimension of $\phi_e$ in the first layer is changed to accommodate the atom coordinates. 

\paragraph{QM9}
On QM9, the EDM and GDMs are trained using EGNNs with $256$ hidden features and $9$ layers. The models are trained for $1100$ epochs, which is around $1.7$ million iterations with a batch size of $64$. The models are saved every $20$ epochs when the validation loss is lower than the previously obtained number. The diffusion process uses $T = 1000$. Training takes approximately $7$ days on a single NVIDIA GeForce GTX 1080Ti GPU. When generating samples the model takes on average $1.7$ seconds per sample on the 1080Ti GPU. These times should not be taken as a fundamental limit of sampling performance, as more efficient samplers can be extracted from diffusion models after training \citep{salimans2022progressive}. For comparison, the E-NF takes 0.54 seconds per sample and G-Schnet 0.03 seconds. The EDM that only models heavy atoms and no hydrogens has the same architecture but is faster to train because it operates over less nodes: it takes about $3.2$ days on a single 1080Ti GPU for $1100$ epochs and converges even earlier to its final performance. 

\paragraph{GEOM-DRUGS}
On GEOM, the EDM and GDMs are trained using EGNNs with $256$ hidden features and $4$ layers. The models are trained for $13$ epochs, which is around $1.2$ million iterations with a batch size of $64$. Training takes approximately $5.5$ days on three NVIDIA RTX A6000 GPUs. The model then takes on average $10.3$ seconds to generate a sample. 

\begin{figure}[h!]
\centering
\includegraphics[width=\textwidth]{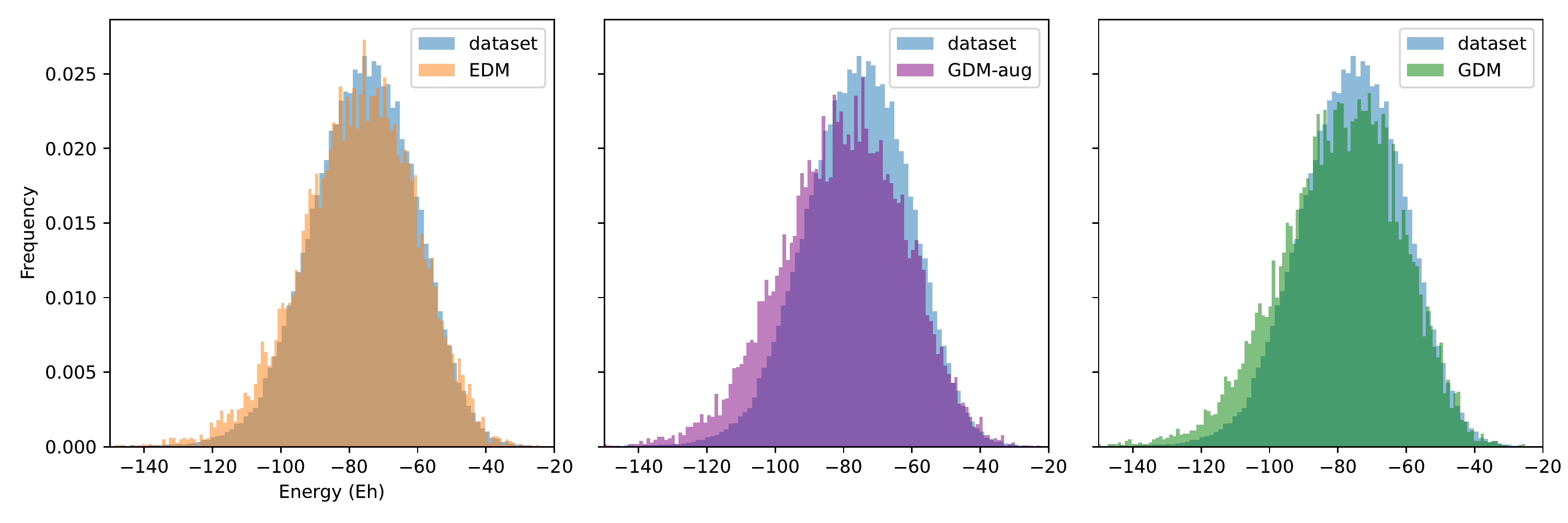}
\caption{Distribution of estimated energies for the molecules generated by all methods trained on GEOM-DRUGS. We observe that EDM captures the dataset distribution well, while other methods tend to produce too many low-energy compounds.}
\label{fig:energy_histograms}
\end{figure}

\paragraph{Limitations of RDKit-based metrics}

Most commonly, unconstrained molecule generation methods are evaluated using RDKit. First, a molecule is built that contains only heavy atoms. Then, RDKit processes this molecule. In particular, it adds hydrogens to each heavy atoms in such a way that the valency of each atom matches its atom type. As a result, invalid molecules mostly appear when an atom has a valency bigger than expected. 

Experimentally, we observed that validity could artificially be increased by reducing the number of bonds. For example, predicting only single bonds was enough to obtain close to $100\%$ of valid molecules on GEOM-DRUGS. Such a change also increases the novelty of the generated molecules, since these molecules typically contain more hydrogens than the training set. On the contrary, our stability metrics directly model hydrogens and cannot be tricked as easily.

Regarding novelty, \citet{vignac2021top} argued that QM9 is the exhaustive enumeration of molecules that satisfy a predefined set of constraints. As a result, a molecule that is novel does not satisfy at least one of these constraints, which means that the algorithm failed to capture some properties of the dataset. Experimentally, we observed that novelty decreases during training, which is in accordance with this observation. The final novelty metrics we obtain are the following:

 \begin{table}[h!]
    \centering
    \caption{Novelty among valid and unique molecules (starting from 10000 molecules) with standard deviations across 3 runs on QM9. Experimentally, we observed that novelty is initially close to $100\%$, and decreases during training. On QM9, it reflects the fact that the algorithm progressively learns to capture the data distribution, which is an exhaustive enumeration under a predefined set of constraints. }
    \scalebox{.9}{
    \begin{tabular}{l r r r}
    \toprule
    Method    &  GDM-aug & EDM (with H) & EDM (no H)\\ 
    Novelty (\%) & 74.6 & 65.7 $\spm{1.3}$& $34.5 \spm{0.9}$ \\\bottomrule
    \end{tabular}}
    \label{tab:qm9-novelty}
\end{table}

\paragraph{Jensen–Shannon divergence between atomic distances}

In addition to the reported metrics in the QM9 experiment, we also report the Jensen-Shannon divergence between histograms of inter-atomic distances. Following the procedure from \cite{satorras2021en_flows}, we produce a histogram of relative distances between all pairs of atoms within each molecule. Then, we compute the Jensen-Shannon divergence between the normalized histograms produced with the generated and the training samples $\mathrm{JS}_{\text{div}}(P_{\mathrm{gen}} || P_{\mathrm{data}})$. The lower the metric, the closer the distribution between generated and training samples will be. We report this metric in the following table for E-NF \citep{satorras2021en_flows}, G-Schnet \citep{gebauer2019symmetry} and EDM.

 \begin{table}[h!]
    \caption{Jensen-Shannon divergence between the normalized histograms of inter-atomic distances within atoms (lower is better). Results are reported for E-NF \citep{satorras2021en_flows}, G-Schnet \citep{gebauer2019symmetry} and our proposed method EDM.}
\begin{center}
\scalebox{.9}{
     \begin{tabular}{l c c c}
    \toprule
     & E-NF & G-Schnet & EDM (Ours)\\
      \midrule
        JS-Div & .0049 & .0027 & \textbf{.0002}\\
     \bottomrule
    \end{tabular}}
\end{center}
\end{table}

\paragraph{Bond distances}
 In order to check the validity and stability of the generated structures, we compute the distance between all pairs of atoms and use these distances to predict the existence of bonds and their order.
Bond distances in Table \ref{tab:bond_typical_distances}, \ref{tab:bond_double_typical_distances} and \ref{tab:bond_triple_typical_distances} are based on typical distances in chemistry\footnote{\url{http://www.wiredchemist.com/chemistry/data/bond_energies_lengths.html}}\footnote{\url{http://chemistry-reference.com/tables/Bond\%20Lengths\%20and\%20Enthalpies.pdf}}. In addition, margins are defined for single, double, triple bonds $m_1, m_2, m_3 = 10, 5, 3$ which were found empirically to describe the QM9 dataset well. If an two atoms have a distance shorter than the typical bond length plus the margin for the respective bond type, the atoms are considered to have a bond between them. The allowed number of bonds per atom are: H: 1, C: 4, N: 3, O: 2, F: 1, B: 3, Al: 3, Si: 4, P: [3, 5], S: 4, Cl: 1, As: 3, Br: 1, I: 1. After all bonds have been created, we say that an atom is stable if its valency is precisely equal to the allowed number of bonds. An entire molecule is considered stable if \textit{all} its atoms are stable. Although this metric does not take into account more atypical distances or aromatic bonds, it is still an extremely important metric as it measures whether the model is positioning the atoms precisely enough. On the QM9 dataset it still considers 95.2\% molecules stable and 99.0\% of atoms stable. For Geom-Drugs the molecules are much larger which introduces more atypical behaviour. Here the atom stability, which is 86.5\%, can still be used since it describes how many atoms satisfy the typical bond length description. However, the molecule stability is 2.8\% on the dataset, which is too low to draw meaningful conclusions.
  
  \begin{table}[!h]
\footnotesize
    \centering
    \caption{Typical bond distances for a single bond.} \label{tab:bond_typical_distances}
    \begin{tabular}{l | r r r r r r r r r r r r r r r r r}
    \toprule
& H & C & O & N & P & S & F & Si & Cl & Br & I & B & As \\ \midrule
H & 74 & 109 & 96 & 101 & 144 & 134 & 92 & 148 & 127 & 141 & 161 & 119 & 152 \\
C & 109 & 154 & 143 & 147 & 184 & 182 & 135 & 185 & 177 & 194 & 214 & - & - \\
O & 96 & 143 & 148 & 140 & 163 & 151 & 142 & 163 & 164 & 172 & 194 & - & - \\
N & 101 & 147 & 140 & 145 & 177 & 168 & 136 & - & 175 & 214 & 222 & - & - \\
P & 144 & 184 & 163 & 177 & 221 & 210 & 156 & - & 203 & 222 & - & - & - \\
S & 134 & 182 & 151 & 168 & 210 & 204 & 158 & 200 & 207 & 225 & 234 & - & - \\
F & 92 & 135 & 142 & 136 & 156 & 158 & 142 & 160 & 166 & 178 & 187 & - & - \\
Si & 148 & 185 & 163 & - & - & 200 & 160 & 233 & 202 & 215 & 243 & - & - \\
Cl & 127 & 177 & 164 & 175 & 203 & 207 & 166 & 202 & 199 & 214 & - & 175 & - \\
Br & 141 & 194 & 172 & 214 & 222 & 225 & 178 & 215 & 214 & 228 & - & - & - \\
I & 161 & 214 & 194 & 222 & - & 234 & 187 & 243 & - & - & 266 & - & - \\
B & 119 & - & - & - & - & - & - & - & 175 & - & - & - & - \\
As & 152 & - & - & - & - & - & - & - & - & - & - & - & - \\
    \bottomrule
    \end{tabular}
\end{table}

\begin{table}[!h]
\centering
\begin{minipage}[t]{.5\textwidth}
  \begin{table}[H]
\footnotesize
    \centering
    \caption{Typical bond distances for a double bond.} \label{tab:bond_double_typical_distances}
    \begin{tabular}{l | r r r r r r r r r r r r r r r r r}
    \toprule
& C & O & N & P & S \\ \midrule
C & 134 & 120 & 129 & - & 160 \\
O & 120 & 121 & 121 & 150 & - \\
N & 129 & 121 & 125 & - & - \\
P & - & 150 & - & - & 186 \\
S & - & - & - & 186 & - \\
    \bottomrule
    \end{tabular}
\end{table}
\end{minipage}
\begin{minipage}[t]{.4\textwidth}
\begin{table}[H]
\footnotesize
    \centering
    \caption{Typical bond distances for a triple bond.} \label{tab:bond_triple_typical_distances}
    \begin{tabular}{l | r r r r r r r r r r r r r r r r r}
    \toprule
& C & O & N \\ \midrule
C & 120 & 113 & 116 \\
O & 113 & - & - \\
N & 116 & - & 110 \\
    \bottomrule
    \end{tabular}
\end{table}
\end{minipage}
\end{table}
  
\paragraph{Limitations of Log-Likelihood}
One of the metrics on which we compare the models is the log-likelihood, which is the negative cross-entropy between the data distribution and the model distribution. For discrete data, such a loss has a direct interpretation: it gives the number of bits required to compress the signal losslessly. For continuous data, no such interpretation exists. Further difficulty is that even though the representation is continuous, the underlying distribution may be discrete. When this happens, the log-likelihood is unbounded and can grow arbitrarily large. For the datasets used in this paper, the positional information (the conformation) is optimized with an iterative process to a local minimum, and thus has a discrete nature. 

For this reason, the log-likelihoods are unbounded and should be treated with caution. They still provide insight into how the model is fitted: Higher log-likelihoods correspond to sharper model distributions on the correct locations, which is an important positive indication that the model is fitted well. However, it can also happen that part of the distribution (suppose for an x-coordinate) is extremely sharp whereas it is very blurred for another part (suppose a y-coordinate). The log-likelihood would still be high because of the x-coordinate, even though the y-coordinate is poorly fitted. Therefore, for this type of data, log-likelihoods should be considered in combination with other metrics such as the atom stability and molecule stability metrics, as done in this paper.

\newpage

\section{Samples from our models}

\begin{figure*}[h!]
\includegraphics[width=1.0\textwidth]{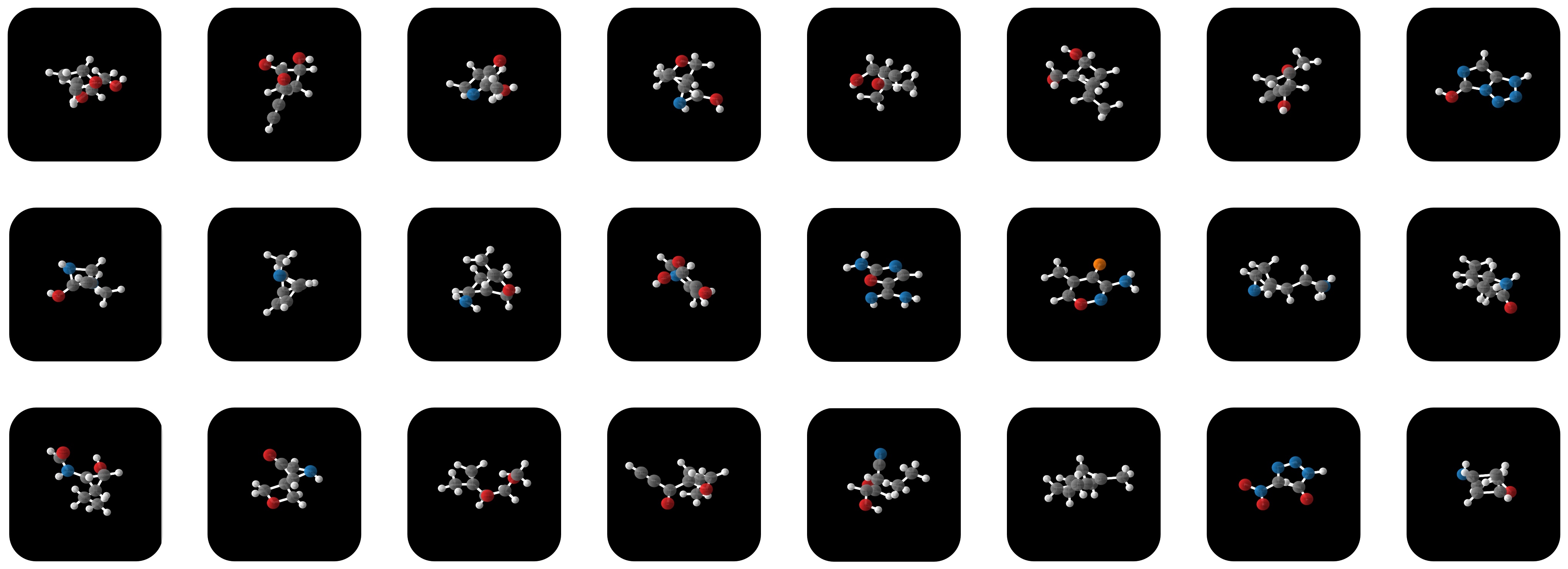}
\centering
\caption{Random samples taken from the EDM trained on QM9.}
\label{fig:generated_molecules_qm9}
\end{figure*}

\begin{figure*}[h!]
\includegraphics[width=1\textwidth]{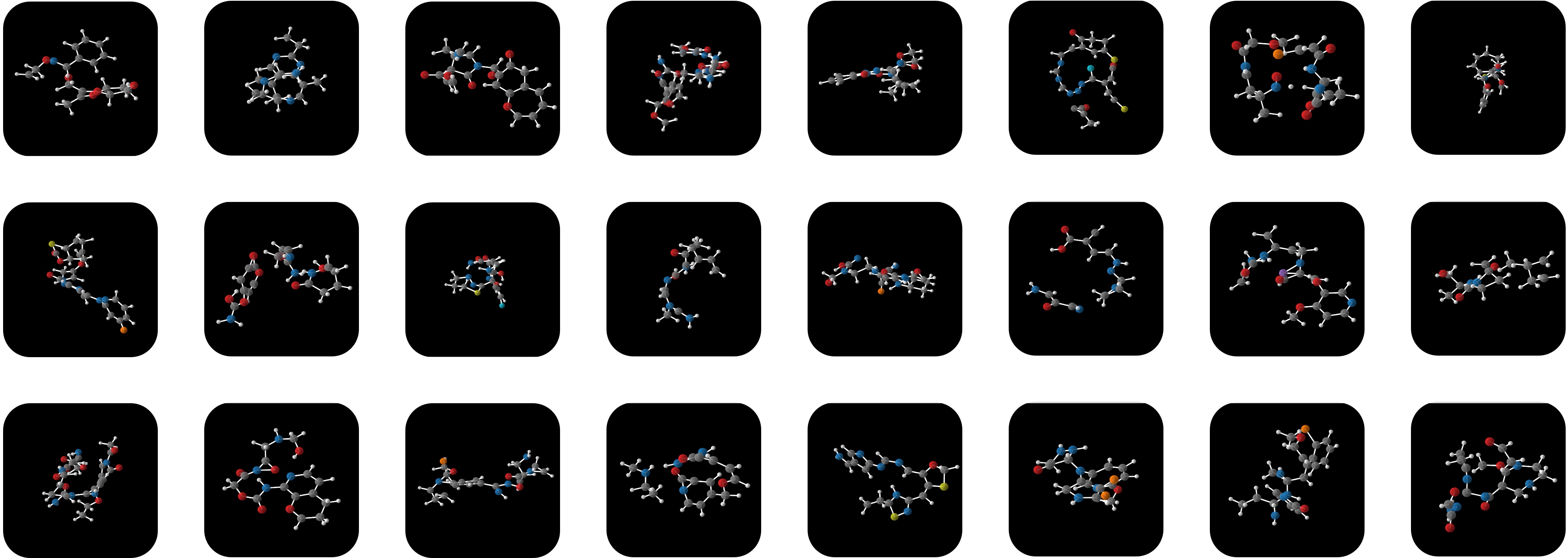}
\centering
\caption{Random samples taken from the EDM trained on geom drugs. While most samples are very realistic, we observe two main failure cases: some molecules that are disconnected, and some that contain long rings. We note that the model does not feature any regularization to prevent these phenomena.
}
\label{fig:generated_molecules_geom_drugs}
\end{figure*}

Additional samples from the model trained on QM9 are depicted in Figure \ref{fig:generated_molecules_qm9} and, and samples from the model trained on GEOM-DRUGS in Figure \ref{fig:generated_molecules_geom_drugs}. These samples are not curated or cherry picked in any way. As a result, their structure may sometimes be difficult to see due to an unfortunate viewing angle.  

The samples from the model trained on the drugs partition of GEOM show impressive large 3D structures. Interestingly, the model is sometimes generating disconnected component, which only happens QM9 models in early training stages. This may indicate that further training and increasing expressitivity of the models may further help the model bring these components together.

\begin{figure*}[h!]
\includegraphics[width=0.88\textwidth]{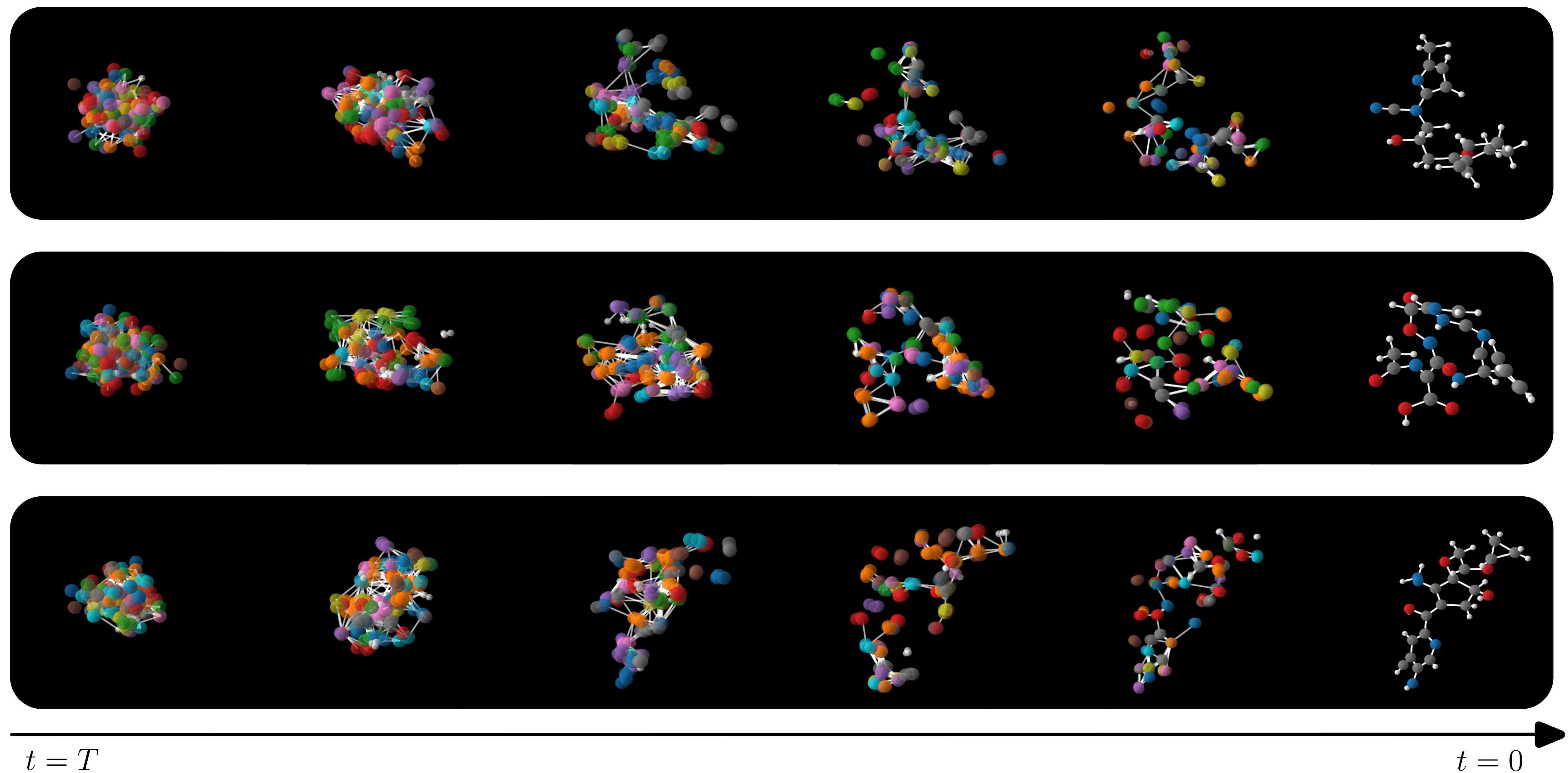}
\centering
\caption{Selection of sampling chains at different steps from a model trained on GEOM-Drugs. The final column shows the resulting sample from the model.}
\label{fig:sampled_t_geom}
\end{figure*}

\newpage
Figure~\ref{fig:sampled_t_geom} depicts the generation of molecules from a model trained on GEOM-Drugs. The model starts at random normal noise at time $t = T = 1000$ and iteratively sample $\vz_{t-1} \sim p(\vz_{t-1} | \vz_t)$ towards $t=0$ to obtain $\vx, \vh$, which is the resulting sample from the model. The atom type part of $\vz_t^{(h)}$ is visualized by taking the argmax of this component.

\paragraph{Ablation on scaling features}
In Table~\ref{tab:appendix_ablation_scaling} a comparison between the standard and proposed scaling is shown. Interestingly, there is quite a large difficulty in performance when measuring atom and molecule stability. From these results, it seems that it is easier to learn a denoising process where the atom type is decided later, when the atom coordinates are already relatively well defined.

 \begin{table}[H]
     \vspace{-8pt}
    \centering
    \caption{Ablation study on the scaling of features of the EDM. Comparing our proposed scaling to no scaling.}\label{tab:appendix_ablation_scaling}
        \scalebox{.9}{
    \begin{tabular}{l r r r r}
    \toprule
     \# Metrics & Scaling & NLL\hspace{0.58cm} & Atom stable (\%) & Mol stable (\%)\\
      \midrule
        EDM (ours) & $[\vx, 1.00 ~\vh^\text{onehot}, 1.0~ \vh^\text{atom charge}]$ & -103.4\hspace{0.58cm} & 95.7\hspace{0.58cm} & 46.9\hspace{0.58cm} \\ 
        EDM (ours) & $[\vx, 0.25 ~\vh^\text{onehot}, 0.1~ \vh^\text{atom charge}]$ & \textbf{-110.7}$\spm{1.5}$ & \textbf{98.7}$\spm{0.1}$ & \textbf{82.0}$\spm{0.4}$ \\ 
        \midrule
        Data &  & & 99.0\hspace{0.58cm} & 95.2\hspace{0.58cm} \\
     \bottomrule
    \end{tabular}}
     \vspace{-15pt}
  \end{table}

\clearpage

\section{Conditional generation} \label{ap:conditional_exp}

\paragraph{Conditional Method}
The specific definition for the loss components $\mathcal{L}_{c,t}$ is given in Equation~\ref{eq:loss_term_epsilon_conditional}. Essentially, a conditioning on a property $c$ is added where relevant. The diffusion process that adds noise is not altered. The generative denoising process is conditioned on $c$ by adding it as input to the neural network $\phi$:
\begin{align}
\small
\begin{split}
\mathcal{L}_{c,t} &= \mathbb{E}_{\vepsilon_t \sim \mathcal{N}_{xh}(0, \rmI)} \big{[}\frac{1}{2} (1 - \mathrm{SNR}(t-1) / \mathrm{SNR}(t)) || \vepsilon_t - \phi(\vz_t, t, c) ||^2\big{]}, \\
    \mathcal{L}_{c,0}^{(h)} &= \log p(\vh | \vz_0^{(h)}) \approx 0, \\
        \mathcal{L}_{c,0}^{(x)} &= \log p(\vx | \vz_0, c) = \mathbb{E}_{\vepsilon^{(x)} \sim \mathcal{N}(0, \rmI)} \Big{[} \log Z^{-1} - \frac{1}{2}||\vepsilon^{(x)} - \phi^{(x)}(\vz_0, 0, c) ||^2 \Big{]}, \\
       \mathcal{L}_{c,\text{base}} &= \mathcal{L}_{\text{base}}  = -\mathrm{KL}(q(\vz_T | \vx, \vh) | p(\vz_T)) \approx 0.
\end{split}
\label{eq:loss_term_epsilon_conditional}
\end{align}

Given a trained conditional model $p(\vx, \vh | c, M)$, we define the generative process by first sampling $c, M \sim p(c, M)$ and then $\vx, \vh \sim p(\vx, \vh | c, M)$. We compute $c, M \sim p(c, M)$ on the training partition as a parametrized two dimensional categorical distribution where we discretize the continuous variable $c$ into small uniformly distributed intervals.

\textbf{Implementation details:} In this conditional experiment, our Equivariant Diffusion Model uses an EGNN with 9 layers, 192 features per hidden layer and SiLU activation functions. We used the Adam optimizer with learning rate $10^{-4}$ and batch size 64. Only atom types (categorical) and positions (continuous) have been modelled but not atom charges. All methods have been trained for $\sim 2000$ epochs while doing early stopping by evaluating the Negative Log Likelihood on the validation partition proposed by \citep{NEURIPS2019_03573b32}. 

Additionaly, the obtained molecule stabilities in the conditional generative case was similar to the the ones obtained in the non-conditional setting. The reported molecule stabilities for each conditioned property evaluated on 10K generated samples are: (80.4\%) $\alpha$, (81.73\%) $\Delta \varepsilon$, (82.81\%) $\varepsilon_{\mathrm{HOMO}}$, (83.6 \%) $\varepsilon_{\mathrm{LUMO}}$, (83.3\%) $\mu$, (81.03 \%) $C_v$.

\textbf{QM9 Properties}

  $\alpha$ Polarizability: Tendency of a molecule to acquire an electric dipole moment when subjected to anexternal electric field.
  
     $\varepsilon_{\mathrm{HOMO}}$: Highest occupied molecular orbital energy.
     
     $\varepsilon_{\mathrm{LUMO}}$: Lowest unoccupied molecular orbital energy.
     
     $\Delta \varepsilon$ Gap: The energy difference between HOMO and LUMO.
     
     $\mu$: Dipole moment.
     
     $C_v$: Heat capacity at 298.15K

\begin{figure*}[t!] 
\includegraphics[width=1.0\textwidth]{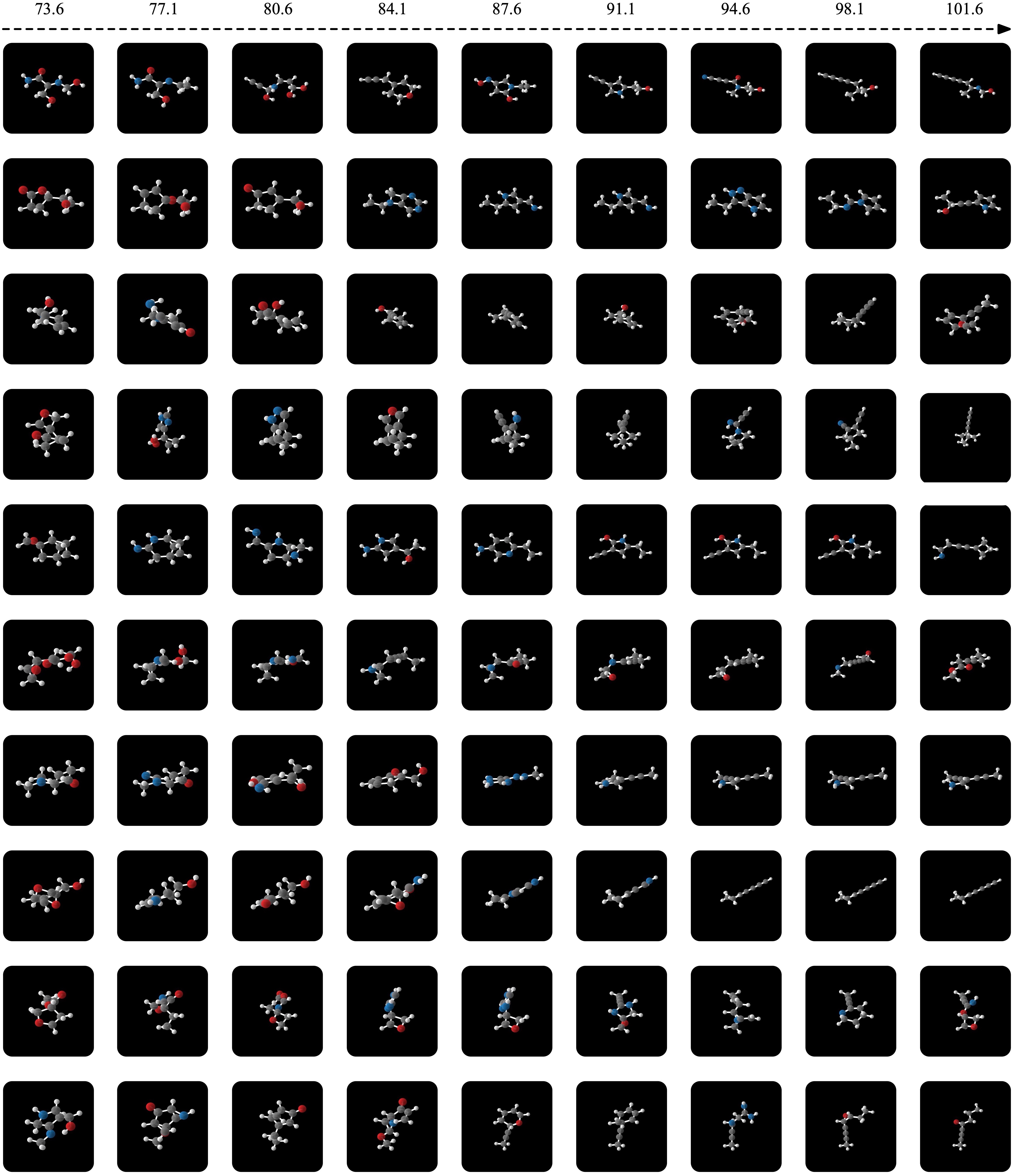}
\centering
\vspace{-15pt}
\caption{Molecules generated by our Conditional EDM when interpolating among different $\alpha$ polarizability values (from left to right). $\alpha$'s are reported on top of the image. All samples within each row have been generated with the same reparametrization noise ${\vepsilon}$.}
\label{fig:conditional_appendix}
\vspace{-15pt}
\end{figure*}

\paragraph{Conditional generation results}
In this Section we sweep over 9 different $\alpha$ values in the range [73.6, 101.6] while keeping the reparametrization noise $\vepsilon$ fixed and the number of nodes $M=19$. We plot 10 randomly selected sweeps in Figure \ref{fig:conditional_appendix} with different reparametrization noises $\vepsilon$ each.  Samples have been generated using our Conditional EDM. We can see that for larger Polarizability values, the atoms are distributed less isotropically encouraing larger dipole moments when an electric field is applied. This behavior is consistent among all reported runs.

\end{document}